\documentclass[lettersize,journal]{IEEEtran}
\usepackage{amsmath}
\usepackage{amsfonts}
\usepackage{amssymb}
\usepackage[ruled,vlined]{algorithm2e}
\usepackage{array}
\usepackage[caption=false,font=normalsize,labelfont=sf,textfont=sf]{subfig}
\usepackage{textcomp}
\usepackage{stfloats}
\usepackage{url}
\usepackage{verbatim}
\usepackage{graphicx}
\usepackage{cite}
\usepackage{booktabs}
\usepackage{multirow}
\usepackage[table]{xcolor}
\usepackage[caption=false,font=normalsize,labelfont=sf,textfont=sf]{subfig}
\usepackage{placeins}
\usepackage{balance}
\begin{document}

\title{Dual-Stream Alignment for Action Segmentation}

\author{Harshala~Gammulle,~\IEEEmembership{Member,~IEEE,}
        Clinton~Fookes,~\IEEEmembership{Senior Member,~IEEE,}
        Sridha~Sridharan,~\IEEEmembership{Life Senior Member,~IEEE,}
        Simon~Denman,~\IEEEmembership{Member,~IEEE.}
        
        % Clinton~Fookes,~\IEEEmembership{Senior Member,~IEEE.}% <-this % stops a space
        \IEEEcompsocitemizethanks{\IEEEcompsocthanksitem H. Gammulle, C. Fookes, S. Sridharan and S. Denman  are with the Signal Processing, Artificial Intelligence and Vision Technologies (SAIVT) Lab, Queensland University of Technology, Brisbane, Australia.\protect\\
E-mail: pranali.gammule@qut.edu.au}
        % <-this % stops a space
}



\maketitle

\begin{abstract}

%------------
Action segmentation is a challenging yet active research area that involves identifying when and where specific actions occur in continuous video streams. Most existing work has focused on single-stream approaches that model the spatio-temporal aspects of frame sequences. However, recent research has shifted toward two-stream methods that learn action-wise features to enhance action segmentation performance.

In this work, we propose the Dual-Stream Alignment Network (DSA\_Net) and investigate the impact of incorporating a second stream of learned action features to guide segmentation by capturing both action and action-transition cues. Communication between the two streams is facilitated by a Temporal Context (TC) block, which fuses complementary information using cross-attention and Quantum-based Action-Guided Modulation (Q-ActGM), enhancing the expressive power of the fused features. To the best of our knowledge, this is the first study to introduce a hybrid quantum-classical machine learning framework for action segmentation. Our primary objective is for the two streams (frame-wise and action-wise) to learn a shared feature space through feature alignment. This is encouraged by the proposed Dual-Stream Alignment Loss, which comprises three components: relational consistency, cross-level contrastive, and cycle-consistency reconstruction losses. 

Following prior work, we evaluate DSA\_Net on several diverse benchmark datasets: GTEA, Breakfast, 50Salads, and EgoProcel. We further demonstrate the effectiveness of each component through extensive ablation studies. Notably, DSA\_Net achieves state-of-the-art performance, significantly outperforming existing methods.

%-------------

\end{abstract}


\section{Introduction}
\IEEEPARstart{R}{eal-world} human actions are inherently continuous and context-dependent. In such settings, understanding not only what an action is but also when it starts and ends is crucial for effective analysis and decision-making. This challenge is addressed through action segmentation, which focuses on recognising the boundaries of actions in time and space.

\begin{figure}[htbp]
        \centering
        	\includegraphics[width=1.0\linewidth]{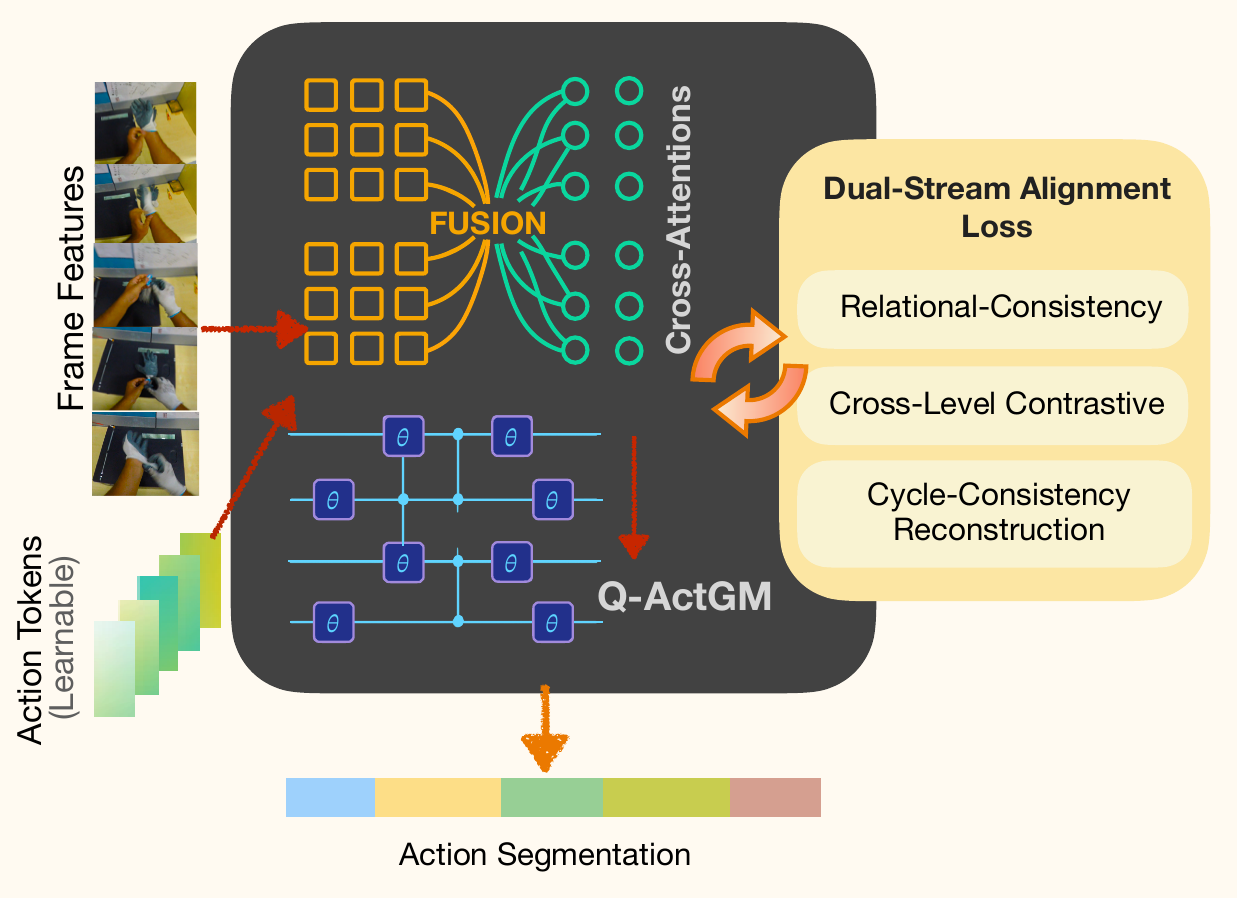}
	\caption{The \textbf{Dual-Stream Alignment Network (DSA\_Net)} supports action segmentation by aligning two streams of features, namely frame features and learnable action tokens, via the Dual-Stream Alignment Loss. Feature fusion across streams is facilitated by the Temporal Context (TC) block, which integrates cross-attention with quantum properties through the proposed Quantum-based Action-Guided Modulation (Q-ActGM) layer.}
	\label{fig:task}
\end{figure}

The continuous and complex nature of human actions has driven significant research interest in action segmentation \cite{farha2019ms, lu2024fact, gammulle2021tmmf}, action localisation \cite{luo2024adaptive, li2024neighbor, liu2022fineaction}, and action detection \cite{liu2022end} over the past decade. While all three approaches aim to recognise and temporally localise human actions, action segmentation provides a comparatively fine-grained understanding by assigning action labels at the frame level, enabling detailed modelling of complex sequential behaviours. Early action segmentation methods were either frame-based \cite{gammulle2019coupled, gammulle2020fine} or sequential models that utilised CRFs \cite{crf1,crf2,crf3} and LSTMs \cite{sadegh2017encouraging}. However, these methods are not easily parallelised and are limited in their ability to capture long-term temporal dependencies. This motivated the introduction of deep temporal models such as temporal convolutional networks (TCNs) \cite{lea2017temporal,farha2019ms}, which have been widely used in both single-stage \cite{lea2017temporal} and multi-stage \cite{farha2019ms,li2020ms} frameworks. More recently, transformer-based methods \cite{chinayi_ASformer,liu2023diffusion,bahrami2023much} have demonstrated superior performance in action segmentation, in particular through a reduction in over-segmentation errors, which cause long actions to be incorrectly broken into multiple smaller actions. This improved accuracy comes at a cost, however, as transformers suffer from high computational complexity as sequence length increases, either due to higher frame rates, longer video sequences, or both.

While the TCN and transformer methods discussed above typically use a single stream of data, that being the input video, Lu et al. \cite{lu2024fact} recently proposed an alternative two-stream approach. In \cite{lu2024fact}, one branch of the network captures conventional frame-wise features to extract low-level details, while the second captures high-level action dependencies via learned action-wise features. Action tokens are learned using a matching loss that ensures each token uniquely encodes an action segment, and a cross-attention mechanism is used to facilitate communication between the two data streams. Overall, this approach was found to enhance action segmentation performance, though there is a significant cost when using cross-attention with long sequences. 

Our proposed approach, the Dual-Stream Alignment Network (DSA\_Net), takes inspiration from \cite{lu2024fact}, as well as recent advancements in state-space models and hybrid quantum methods. Similar to \cite{lu2024fact}, DSA\_Net employs two data streams (frame-wise and action-wise) for long-term temporal action segmentation. In contrast to \cite{lu2024fact} and other recent approaches that employ convolution-based \cite{lea2017temporal, farha2019ms, li2020ms} or transformer-based \cite{chinayi_ASformer, liu2023diffusion, bahrami2023much} encoders, our method leverages state-space models \cite{mamba2}. We formulate a simple yet effective Temporal Encoder (TE) block based on a state-space representation that employs a selective gating mechanism to filter out irrelevant features. We demonstrate the effectiveness of this approach for temporal modelling in action segmentation. 

Following \cite{lu2024fact}, we define the action-wise features as learnable action tokens; however, unlike \cite{lu2024fact}, we learn these tokens via a dual-stream alignment loss. This loss is itself composed of three components, designed to encourage the deep distillation of complementary information and promote alignment between the two streams. A relational consistency component encourages pairwise similarity between frames to be reflected in the action tokens. A cross-level consistency term seeks to align action-token embeddings with the frame embeddings they attend to, while separating them from less relevant frames. Finally, a cycle-consistency reconstruction term encourages the learning of cross-stream relationships by ensuring that action tokens can be used to reconstruct frame features and vice versa. Together, these three components encourage the learning of more discriminative segment-wise representations, thereby better capturing action semantics.

To enhance the sharing of information across the two data streams, we incorporate ideas from quantum machine learning and the feature-wise linear modulation approach of \cite{perez2018film}. Similar to \cite{lu2024fact}, we also utilise cross-attention to facilitate cross-branch communication; however, drawing inspiration from \cite{perez2018film}, we enhance cross-stream communication by adopting feature-wise linear modulation for cross-stream information fusion. Specifically, we perform cross-stream attention-based fusion while modulating frame features based on action-context embeddings. This is further enhanced by estimating feature-modulation parameters using a parameterised quantum circuit (PQC), leveraging the quantum properties of superposition and entanglement to improve feature expressiveness. To the best of our knowledge, this is the first work to explore quantum-based feature modulation and the first to introduce a hybrid quantum-classical approach for action segmentation.

A summary of our contributions is listed below:
\begin{itemize}
    \item We propose a novel action segmentation framework, DSA\_Net, which aligns frame-wise and action-wise feature streams to learn richer representations than single-stream counterparts.
    \item We introduce the Temporal Context (TC) block, designed to fuse information from the two streams using cross-attention. To further enhance the expressiveness of the fused features, we incorporate quantum-based estimation of modulation parameters through the proposed Q-ActGM layer. To the best of our knowledge, this is the first work to apply a hybrid quantum-classical machine learning formulation to the task of action segmentation.
    \item To encourage alignment between frame and action features, we propose a dual-stream alignment loss composed of three components: relational consistency, cross-level contrastive loss, and cycle-consistency reconstruction. Together, these losses enable the network to distil complementary information more effectively from each stream.
    \item We conduct extensive evaluations on four diverse benchmark datasets, complemented by ablation studies, to demonstrate the effectiveness of our contributions.
\end{itemize}

\section{Related Work}

Action segmentation has become a core task in video understanding, focusing on identifying distinct human actions and their transitions within a video sequence. Capturing both spatial and temporal features is therefore critical for accurately recognising actions and their boundaries.

% temporal models
Considering the continuous and sequential nature of the action segmentation task, temporal models have been widely investigated. Early works focused on classical models such as Hidden Markov Models (HMMs) \cite{kuehne2014language, niebles2010modeling}, Conditional Random Fields (CRFs) \cite{sener2015two, wang2010hidden}, and grammar-based approaches \cite{vo2014stochastic, pirsiavash2014parsing}, which aimed to learn the hierarchical structure of actions or activities. However, these traditional methods relied on handcrafted features, making them incapable of learning complex spatio-temporal patterns and unable to capture long-term temporal dependencies. This limitation motivated the adoption of deep temporal models, with the earliest methods based on recurrent networks such as RNNs \cite{rumelhart1985learning}, LSTMs \cite{hochreiter1997long}, and GRUs \cite{cho2014learning}. Yet, due to their limited ability to model long-range temporal patterns, recurrent models struggled to achieve high performance on videos of longer durations.

Subsequently, attention shifted toward Temporal Convolutional Networks (TCNs). Since the introduction of the initial TCN methods \cite{lea2017temporal}, numerous TCN-based approaches have been developed \cite{farha2019ms, li2020ms, chen2020action, lea2016segmental, lei2018temporal, gammulle2021tmmf} to better capture temporal patterns and reduce over-segmentation errors. However, the effectiveness of TCNs in modelling temporal relations is strongly dependent on the size of their receptive fields.

Following the emergence of transformer architectures in other domains, ASFormer \cite{yi2021asformer} introduced a transformer encoder–decoder for action segmentation. Since then, several transformer-based approaches \cite{aziere2022multistage, bahrami2023much, liu2023diffusion, lu2024fact} have demonstrated significant improvements in segmentation performance. These performance gains, however, are tempered by the quadratic growth in computational complexity that transformers suffer with respect to video sequence length. Recently, new methods \cite{chaudhuri2024simba, sinha2025ms} have leveraged advances in state-space models, such as Mamba \cite{mamba2}, to capture long-range temporal patterns for video-based action understanding. Although these developments are promising, the application of Mamba-based architectures to action segmentation remains in its early stages.

% two-stream approaches and feature fusion and losses that encourage feature fusion
Most of the methods discussed so far rely primarily on frame-wise features. In contrast, \cite{lu2024fact} introduced a two-branch approach with a second stream that learns action-segment-wise features. Prior to this, several action segmentation methods explored the use of action relations \cite{ahn2021refining, behrmann2022unified, jiang2023video, huang2020improving}. However, in these methods, action features were learned only after the initial frame features and predictions had been obtained. In \cite{lu2024fact}, the authors demonstrated that jointly learning action-wise features and leveraging them to refine frame-wise features through fusion is more effective for improving action segmentation performance. Their approach employed a bidirectional cross-attention mechanism and learned action tokens via a matching loss. By contrast, our proposed framework achieves this using a dual-stream alignment loss together with a Temporal Context block, which enables interactions between the streams through attention and feature modulation.

% hybrid methods and hybrid methods for fusion

The fusion of features from multiple data streams is inherently challenging, and numerous methods have been proposed to enable dynamic and effective integration of information. In \cite{perez2018film}, the authors introduced the feature-wise linear modulation layer, demonstrating its effectiveness in visual question answering, where linguistic inputs modulated visual feature representations in neural networks. Following this work, subsequent studies have explored feature modulation as a mechanism for multimodal fusion \cite{brousmiche2022multimodal, ghadiya2024cross}. These works have primarily focused on directly fusing modalities such as audio and visual data for tasks like event prediction and anomaly detection. Our proposed method differs from these earlier works in that the second feature stream (i.e., action tokens) is learned jointly during training, while also supporting action-segment–aware learning of frame-wise features.

To further enhance the fusion process, we consider the use of hybrid classical–quantum machine learning, an emerging area that has shown promising recent progress \cite{islam2022hybrid, qu2023qnmf, tiwari2024quantum}. Various hybrid approaches have been introduced across domains such as healthcare \cite{qu2023qnmf}, surveillance \cite{majumder2021hybrid, khan2023hybrid}, and cybersecurity \cite{islam2022hybrid}. In the context of fusion-based methods, however, only a limited number of hybrid approaches have been proposed \cite{tiwari2024quantum, qu2023qnmf}. In these cases, fusion has typically relied on simple concatenation (either direct or attention-based), or on straightforward PQC, which limits the depth of interaction between feature streams. In contrast, our approach introduces quantum-based feature modulation, a more expressive mechanism that conditions and reshapes feature representations, enabling deeper cross-stream integration. This richer fusion leads to significant gains in action segmentation performance, highlighting the potential of quantum-enhanced modelling in video understanding.

\section{Methods}
\label{sec:methods}

\subsection{Overview}

\begin{figure*}[!t]
    \centering
    \includegraphics[width=0.69\linewidth]{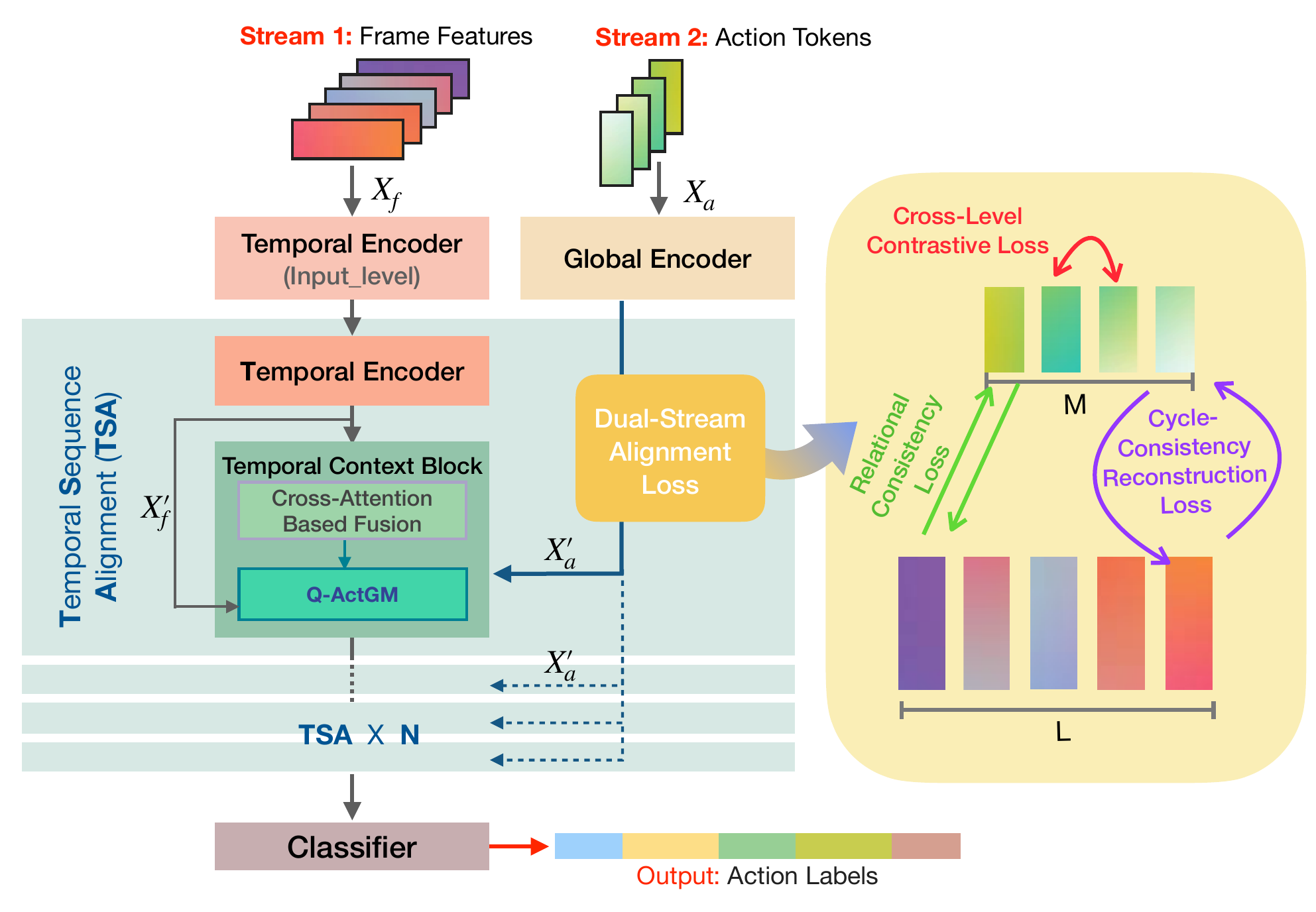}
	\caption{Overview of the proposed DSA\_Net: The model maintains two streams of features, frame features and action tokens, while modelling their temporal dynamics through Temporal Encoders (TE) and a Global Encoder (GE), respectively. Feature fusion is performed via the Temporal Context (TC) block, within the Temporal Sequence Alignment (TSA) block. The TC block integrates a cross-attention mechanism with the proposed Quantum-based Action-Guided Modulation (Q-ActGM), which introduces quantum properties to enhance expressive power. Feature alignment is encouraged through the proposed Dual-Stream Alignment loss.}
	\label{fig:DSA_net}
\end{figure*}

In action segmentation, the goal is to provide dense, frame-wise predictions that identify both the action classes and their temporal boundaries. Given an input video $X_{1:T} = (X_{1}, \dots, X_{T})$ with $T$ frames, the goal is to predict the frame-wise class labels $Y_{1:T}={Y_1, \dots, Y_{T}}$. As these videos often contain a large number of frames and multiple action segments and transitions which flow in a continuous manner, it is crucial to capture long-term temporal cues. 

% In this work, we introduce the Dual-Stream Alignment Network (DSA\_Net) for action segmentation. An overview of our proposed DSA\_Net is shown in Figure \ref{fig:DSA_net}. Inspired by the advancements of the recent two-stream approach of Lu et al. \cite{lu2024fact}, we adopt a similar strategy when formulating DSA\_Net, maintaining two distinct information streams: a frame stream and an action stream. While maintaining these two streams, we aim to capture deep spatio-temporal cues from both streams to support the final action segmentation task through the dual-stream alignment formulation. We discuss the main components of the proposed DSA\_Net in detail in the following sections. 

In this work, we introduce the Dual-Stream Alignment Network (DSA\_Net) for action segmentation. An overview of the proposed framework is presented in Figure \ref{fig:DSA_net}. Inspired by the recent two-stream approach of Lu et al. \cite{lu2024fact}, we adopt a similar strategy in DSA\_Net, maintaining two distinct information streams: a frame stream and an action stream. Through this dual-stream alignment formulation, we aim to capture deep spatio-temporal cues from both streams to support the final action segmentation task. The main components of DSA\_Net are discussed in detail in the following sections.

\subsection{DSA\_Net Architecture}
\textbf{Inputs:} As DSA\_Net is a two-stream network, we maintain frame-wise and action-wise input features. Frame-wise inputs for DSA\_Net are pre-extracted frame-wise spatio-temporal features denoted by $X_f \in\mathbb{R}^{L\times d_f}$, while the action-wise features are denoted by $X_a\in\mathbb{R}^{M\times d_a}$. Here, $L$ and $M$ are the lengths of the frame and action feature sequences, respectively, while $d_f$ and $d_a$ represent their corresponding feature dimensions. Similar to \cite{lu2024fact}, our action features $X_a$ are defined as learnable action tokens and initialised as $X_a=0$. In action segmentation, temporal order plays a critical role in understanding actions and their transitions. Therefore, we incorporate a positional encoding to obtain temporally aware frame-wise and action-wise features. However, to maintain simplicity, we omit the inclusion of the positional encoding from the following equations.

Our proposed DSA\_Net consists of a Global Encoder block that learns the spatio-temporal representations of the action-wise features, and a Temporal Sequence Alignment (TSA) module that models spatio-temporal features from the frame-wise feature sequence, while also facilitating alignment between the two streams (i.e., the frame stream and the action stream) to support action segmentation. The following sections discuss these components in detail.

\subsubsection{Global Encoder (GE) Block}

The Global Encoder (GE) aims to capture temporal patterns across the learned action tokens to support the temporal alignment of action-related cues that flow from the two input streams. Given its lightweight architecture and effective temporal modelling, we adopt the single-stage network proposed in \cite{farha2019ms} as our Global Encoder (GE) block,
\begin{equation}
    X'_a = f_{GE}(X_a),
\end{equation}
where $X'_a\in\mathbb{R}^{M\times{d_{a_t}}}$. Here, $d_{a_t}$ is the output feature dimension of the GE block.  

\subsubsection{Temporal Sequence Alignment (TSA) Module}

We design the TSA module to model spatio-temporal features and to learn the alignment between the two streams. The TSA module is composed of Temporal Encoder (TE) and Temporal Context (TC) Blocks, which are described below.  

% \noindent\textbf{Temporal Encoder (TE)} 
\paragraph{Temporal Encoder (TE) Block} The TE block aims to perform temporal modelling to capture long-term temporal cues from the frame-wise feature sequences. The formulation of the Temporal Encoder is inspired by the Mamba architecture \cite{mamba2}, and we utilise state-spaces together with a state-space selection mechanism in a simplified manner as described below.  

Following the Temporal Encoder block, the frame features $X_f$ are projected to an expanded feature space through a linear transformation,
\begin{equation}
    X_{proj} = W_f X_f + b_f,
\end{equation}
where $W_f$ and $b_f$ are a learnable projection matrix and bias term, respectively. This transformation to a higher-dimensional space allows the model to capture more detailed features from the input feature sequence, allowing later steps to better model temporal dynamics and feature dependencies. After the feature projection, state space transformations are applied,
 \begin{equation}
     S = \tanh(W_A X_{proj}^{\intercal}) ,
 \end{equation}
 
 \begin{equation}
     S' = GELU(W_{B}S + W_C),
 \end{equation}
where $W_A, W_B, W_C$ are learnable parameters, and the tanh and GELU activations introduce non-linearities and aid in stabilising the training process by smoothing the activation values. Then, a gating mechanism is applied to select only the relevant state spaces. The gate is defined using a sigmoid function ($\sigma$),
\begin{equation}
   G =\sigma(W_gX_f + b_g), 
\end{equation}
where $W_g, b_g$ are the learnable gating matrix and the bias term, respectively. Once the gate is defined, the gated state is computed,
\begin{equation}
    \overline{S} = [S']^{\intercal} \odot G,
\end{equation}
where $\odot$ represents the element-wise multiplication. The output of the TE block is obtained by projecting the gated state ($i.e. \overline{S}$) back to the original feature dimension,
\begin{equation}
    X'_f = W_{out}\overline{S} + b_{out},
\end{equation}
where $W_{out}$ is a learnable transformation matrix while $b_{out}$ is a bias term. 
% The output $x'_f \in\mathbb{R}^{L\times d_{f_t}}$.

The TE first block that is applied directly to the frame features at the input level (see Figure \ref{fig:DSA_net}) serves as an encoder to compress the input features to a lower-dimensional feature space. Subsequent TE blocks within the TSA module, however, have identical input and output dimensions, $d_h$, allowing multiple TSA blocks to be stacked.

\paragraph{Temporal Context (TC) Block} The TC block merges information that flows through the two streams (frame and action streams), facilitating the dual-stream alignment loss (discussed in sec.\ref{sec:losses}). 
% To allow the comparison between the frame-wise and action-wise information that is passed through TE and the GE blocks, respectively, 

The TC block uses the frame-wise and action-wise features that are passed from the preceding TE and GE blocks, and applies linear projections through attention. $Q$, $K$ and $V$ values are calculated as follows,

\begin{equation}
    Q = X'_f W_q + b_q,
\end{equation}
\begin{equation}
    K = X'_a W_k + b_k,
\end{equation}
\begin{equation}
    V = X'_a W_v + b_v,
\end{equation}

where $W_q$, $W_k$ and $W_v$, and $b_q$, $b_k$ and $b_v$ are the weights and the biases respectively.
 % % where $Q\in\mathbb{R}^{L\times{d_h}}$ and $K, V \in\mathbb{R}^{M\times{d_h}}$.
 % where $W_q\in\mathbb{R}^{d\times{f_t}}$, $W_k, W_v\in\mathbb{R}^{d\times{a_t}}$, $b_q, b_k, b_v \in\mathbb{R}^{H}$.

$Q$ values are derived from the frame-wise features, $X'_f$; while $K$ and $V$ values are derived from action token features, $X'_a$. Cross-attention is then applied as follows,

 \begin{equation}
     A = f_{softmax}(Q K^T/\sqrt{d_h}),
    \label{eq:eq11}
 \end{equation}

\begin{equation}
    A' = A V,
\end{equation}

where $f_{softmax}$ refers to the softmax activation function.
Once $A'$ is calculated, modulation parameters $\gamma$ and $\beta$ are computed to perform feature-wise affine transformations. To determine modulation parameters, we introduce a Quantum-based Action-Guided Modulation (Q-ActGM) layer inspired by \cite{perez2018film}. However, we adapt the feature modulation to achieve cross-stream fusion instead of the modulation of a single feature stream. Our proposed Q-ActGM module is represented as a parameterised quantum circuit. The quantum-based formulation allows the model to explore quantum-enhanced representations for conditioning feature maps. In Algorithm \ref{alg:psudocode}, we illustrate the functionality of the proposed Q-ActGM layer, and we further discuss the step-by-step process below. 

To prepare the embedding, $A'$, for Q-ActGM, we use a linear projection to match the embedding to the quantum input size, 
\begin{equation}
    X_q = f_{linear}(A'_t),
\end{equation}
where $X_q \in\mathbb{R}^{L\times{n_q}}$, and $n_q$ refers to the number of qubits.  

Then the computed $X_q$ is used to encode the data into a parameterised quantum circuit. Each vector at each timestep $X^i_q$ (where $i=1,.., L$) is used as the input to the quantum circuit,

\begin{equation}
    Z^i = U_\theta(X^i_q),
    \label{eq:zi}
\end{equation}

where $U_\theta$ is a parameterised quantum circuit (PQC) that consists of data encoding operations and a trainable entangling layer. Through applying a PQC, we add non-linearity and entanglement properties to the modulation. In Equation \ref{eq:zi}, $Z^i$ returns expectation values of quantum observables (e.g. Pauli-Z), one per qubit. Once the expectation values are calculated, the quantum outputs are mapped back to the classical modulation parameters using a classical linear projection.

\begin{equation}
    [\gamma^i, \beta^i] = f'_{linear}(Z^i).
\end{equation}

As the above calculations are performed at each timestep separately, we next reshape the modulation parameters to match the temporal structure. Let $\gamma'$ and $\beta'$ be the reshaped modulation parameters. Then the frame feature modulation is performed using,% Then these are reshaped to match the temporal structure of the modulation parameters, where $\gamma'$, $\beta'$ $\in\mathbb{R}^{L\times d'_f}$. The next step is to modulate the frame features,
\begin{equation}
    X^*_f = \gamma' \odot X'_f + \beta', 
\end{equation}
to obtain the TSA module output.

Our DSA\_Net is formulated by stacking $N$ TSA blocks to map the spatio-temporal cues of the feature streams. By stacking TSA blocks, we achieve refinement of features with each successive block.
% , instead of the refinement of the predictions done in previous approaches \cite{farha2019ms,lu2024fact}.     

The output of the final TSA block is passed through a classification model to obtain a frame-wise action label sequence,
\begin{equation}
    Y_{out} = f_{classify}(X^*_{f, N}),
\end{equation}
where $X^*_{f, N}$ is the output of the $N^{th}$ TSA block.

%----------------------------
\begin{algorithm}[htbp]
\caption{Quantum-based Action-Guided Modulation (Q-ActGM) Circuit}
\label{alg:quantum-actgm}
\KwIn{Classical feature vector $X'_f$, quantum weights $\Theta \in \mathbb{R}^{n_{ql} \times n_q \times n_p}$, number of qubits $n_q$, number of layers $n_{ql}$, number of parameters per qubit $n_p$}
\KwOut{Quantum output vector $\vec{Z} \in \mathbb{R}^{n_q}$}

Initialize quantum device $\mathcal{D}$ with $n_q$ qubits\;

\BlankLine
\textbf{Embedding Rotation:} \\
\For{$i \gets 1$ \KwTo $n_q$}{
    Apply $RY(X'_{f,i})$ on qubit $i$\;
}

\BlankLine
\textbf{Parameterized Strongly Entangling Layers:} \\
\For{$\ell \gets 1$ \KwTo $n_{ql}$}{
    \For{$i \gets 1$ \KwTo $n_q$}{
        Apply $RY(\Theta_{\ell,i,0})$ on qubit $i$\;
        Apply $RZ(\Theta_{\ell,i,1})$ on qubit $i$\;
        Apply $RX(\Theta_{\ell,i,2})$ on qubit $i$\;
    }
    \For{$i \gets 1$ \KwTo $n_q - 1$}{
        Apply $CNOT$ from qubit $i$ to $i+1$\;
    }
    % Optionally: Apply $CNOT$ from qubit $n_q$ to qubit $1$ (ring entanglement)\;
}

\BlankLine
\textbf{Measurement:} \\
\For{$i \gets 1$ \KwTo $n_q$}{
    Measure expectation value of $Z$ on qubit $i$ and store in $\vec{Z}_i$\;
}

\Return $\vec{Z}$
\label{alg:psudocode}
\end{algorithm}
%-------------------

\paragraph{Q-ActGM Circuit Operation}
\label{para:q_circuit}

We present the operation of the Q-ActGM circuit in Algorithm \ref{alg:psudocode}. We first initialise the quantum device and set up the quantum circuit with $n_q$ qubits. We then apply an embedding rotation, where for each qubit $i$, a rotation around the Y-axis (through an RY gate \cite{schuld2020circuit}) is applied based on the input feature $X'_{f,i}$. Superpositions are introduced at this step, where superposition refers to the qubit's ability to exist in a combination of both 0 and 1 states simultaneously (whereas classical bits can exist only as either 0 or 1). Through this process, the classical data is embedded into a quantum state. 

After rotation, we establish entanglement. To achieve this, we employ a parameterised strongly entangling layer consisting of $n{ql}$ layers. In each layer, three parameterised rotations (around the Y, Z, and X axes) are applied to each qubit, followed by CNOT (Controlled-NOT) gates~\cite{schuld2019quantum} between adjacent qubits to entangle them. The final output is stored in the output vector $\vec{Z}\in \mathbb{R}^{n_q}$, which represents a quantum-enhanced feature vector derived from the transformed input. 

% In Figure \ref{fig:quantum-circuits}, we visualise our quantum circuit with three different configurations by setting $n_q$ to 2, 3 and 4, while keeping $n_{ql}=3$ constant. 

% Optionally, CNOTs are applied from $(n_q)^{th}$ qubit to the $1^{st}$ qubit, forming ring entanglement.

% \begin{figure*}[t!]
%     \centering
%     \begin{subfigure}
%         \centering
%         \includegraphics[width=\textwidth]{Figures/q2l3.png}
%         % \caption{}
%         \label{fig:subfig-a}
%     \end{subfigure}
%      \hfill
%     \begin{subfigure}
%         \centering
%         \includegraphics[width=\textwidth]{Figures/q3l3.png}
%         % \caption{}
%         \label{fig:subfig-b}
%     \end{subfigure}
%     \hfill
%     \begin{subfigure}
%         \centering
%         \includegraphics[width=\textwidth]{Figures/q4l3.png}
%         % \caption{}
%         \label{fig:subfig-b}
%     \end{subfigure}
    
%     \caption{Quantum circuits with different configurations: \( n_q = 2 \), \( n_{ql} = 3 \) (top); \( n_q = 3 \), \( n_{ql} = 3 \) (middle);  \( n_q = 4 \), \( n_{ql} = 3 \) (bottom).}
%     \label{fig:quantum-circuits}
% \end{figure*}

\subsection{Dual-Stream Alignment Loss Formulation}
\label{sec:losses}
Our dual-stream alignment loss comprises 3 components: a relational consistency loss, a cross-level contrastive loss, and a cycle consistency reconstruction loss. These losses are designed to distil information between the frame and the action streams, and support their alignment.

The \textbf{Relational Consistency Loss ($L_{rel}$)} encourages pair-wise similarity between frames to be reflected as a similarity in the structure of the action tokens. Let $h_f$ and $h_a$ be the flattened feature output from the $N^{th}$ TSA block (i.e. the final TSA block) and the flattened action tokens, respectively. Here, $h_f \in\mathbb{R}^{L\times d_h}$ and $h_a \in\mathbb{R}^{M\times d_a}$, where $d_h$ and $d_a$ refer to the hidden dimension of the TSA output feature and the GE blocks, respectively. To calculate $L_{rel}$, we first compute the gram similarity matrices per sample, considering frame and action streams,

\begin{equation}
    G^f = h_f[h_f]^\intercal,
\end{equation}

\begin{equation}
    G^a = h_a[h_a]^\intercal,
\end{equation}

where $G^f$ and $G^a$ matrices are of shape $L\times L$ and $M\times M$, respectively. Once the similarity matrices are calculated, we downsample $G^f$ to $M\times M$ through average pooling each non-overlapping block of size [L/M]. Let the downsampled $G^f$ be denoted $\bar G^f$, then $L_{rel}$ is calculated by normalising $\bar G^f$ and $G^a$ by their Frobenius norms, and computing the difference between them,
\begin{equation}
    \mathcal{L}_{\text{rel}} = \left\lVert \frac{\bar{G}^f}{\lVert \bar{G}^f \rVert_F} - \frac{G^a}{\lVert G^a \rVert_F} \right\rVert_F^2.
\end{equation}

This ensures that the similarity between frames is mirrored in the action tokens stream, aligning their geometric structure. 

% The \textbf{Cross-Level Contrastive Loss ($L_{clc}$)} pulls action token embeddings close to the frame embeddings they attend to, while pushing them away from other frames. Let $a_{n,t}$ be the attention weight between frame $t$ and token $n$ (computed in Eq. \ref{eq:eq11}). For each token $n$, we form its positive set of frames, \{$(t,a_{n, t})$\} and use $a_{n,t}$ as soft weights for our contrastive loss. Then, inspired by \cite{infonc}, we define a temperature-scaled cross-level contrastive loss,
% \begin{equation}
% \mathcal{L}_{\text{clc}} = - \sum_{n=1}^{M} \sum_{t=1}^{L} a_{n,t} \log \frac{\exp\left(\text{sim}(h_n^a, h_t^f)/\tau\right)}{\sum_{t'=1}^{L} \exp\left(\text{sim}(h_n^a, h_{t'}^f)/\tau\right)},
% \end{equation}
% where $\text{sim}(u, v) = \frac{u^\top v}{\lVert u \rVert \lVert v \rVert}$ and $\tau$ is a temperature parameter. This encourages each token $h^a_n$ to be close to its attended frames and distinct from the rest dynamically per batch.

The \textbf{Cross-Level Contrastive Loss ($L_{clc}$)} is designed to align action token embeddings with the frame embeddings that they attend to, while separating them from less relevant frames. 
Let $a_{n,t}$ denote the attention weight between action token $n$ and frame $t$ (defined in Eq.~\ref{eq:eq11}). For each token $n$, we regard all frames as potential matches, but we weight their contribution using the attention scores $a_{n,t}$. In other words, the attended frames form a soft positive set for token $n$, while the remaining frames act as negatives.

Following the InfoNCE formulation \cite{infonc}, we define a temperature-scaled contrastive objective:

\begin{equation}
\mathcal{L}_{\text{clc}} = - \sum_{n=1}^{M} \sum_{t=1}^{L} a_{n,t} \log \frac{\exp\left(\text{sim}(h_n^a, h_t^f)/\tau\right)}{\sum_{t'=1}^{L} \exp\left(\text{sim}(h_n^a, h_{t'}^f)/\tau\right)},
\end{equation}

where $\text{sim}(u, v) = \frac{u^\top v}{\lVert u \rVert \lVert v \rVert}$ is the cosine similarity and $\tau$ is a temperature parameter. This loss encourages each action token, $h_n^a$, to stay close to the frame embeddings it attends to (proportional to $a_{n,t}$), while being pushed away from other frames. The soft weighting allows the model to dynamically determine which frames act as positives versus negatives within each batch.

The \textbf{Cycle-consistency Reconstruction Loss ($L_{cyc}$)} encourages the learning of cross-stream relationships. Specifically, we ensure that action tokens can be used to reconstruct frame features and vice versa, as follows:
\begin{itemize}
    \item (token $\rightarrow$ frame) reconstruction: Let $P^a$ be the token predicted class logits (pre-softmax) and $a_{t,n} \in \mathbb{R}^{L\times M}$ be the frame to action token attention (computed in Eq.\ref{eq:eq11}). Then the frame logits can be reconstructed using, 
    \begin{equation}
        \bar P^f_t= \sum_{n=1}^{M} a_{t,n} P^a_n.
    \end{equation}
    The token to frame reconstruction loss can then be defined using the cross-entropy between the ground truth frame labels ($y^f_t$),
    \begin{equation}
        L^f_{cyc} = \frac{1}{L}\sum_{t} CE(\bar P^f_t, y^f_t).
    \end{equation}
    \item (frame $\rightarrow$ token) reconstruction: Let $\rho$ be the token to frame attention computed by swapping the K and Q vectors in Eq. \ref{eq:eq11} (reversing the direction of action) and $P^f$ be the frame stream based class logits. Following a similar approach, we can calculate $\bar P^a_t$ and the cross‐entropy with the token‐level pseudo labels, which are derived from frames. This can be defined as,
    \begin{equation}
        \bar P^a_t= \sum_{n=1}^{L} \rho_{t,n} P^f_n,
    \end{equation}

    \begin{equation}
        L^a_{cyc} = \frac{1}{M}\sum_{t} CE(\bar P^a_t, y^a_t).
    \end{equation}
\end{itemize}

Then $L_{cyc}$ can be computed,

\begin{equation}
    L_{cyc} = L^a_{cyc} + L^f_{cyc}.
\end{equation}

Once all three loss components are calculated, they are combined to compute the final dual-stream alignment loss,

\begin{equation}
    L_{tot} = L_{ce_f} + L_{ce_a} + L_{rel} + L_{clc} + L_{cyc},
\end{equation}

where $L_{ce_f}$ and $L_{ce_a}$ are the frame-based and action token-based cross-entropy losses calculated using labels predicted from each stream.

\section{Experiments}

Following the state-of-the-art methods, we evaluate our DSA\_Net on four diverse datasets: Breakfast \cite{kuehne2014language}, GTEA \cite{fathi2011learning}, 50 Salads \cite{stein2013combining}, and EgoProceL \cite{bansal2022my}. We compare the obtained results with the current state-of-the-art on each dataset and also perform ablation experiments to demonstrate the contributions of the important components of the proposed DSA\_Net.  

\subsection{Datasets} 
\textbf{GTEA} is based on 7 daily kitchen tasks recorded through a head-mounted GoPro video camera. The dataset contains 28 videos with a total of 4 hours of recordings. This dataset includes 11 distinct actions, and each video contains an average of 33 segments. The \textbf{Breakfast} dataset consists of scenarios where people prepare breakfast, and is captured through a static RGB camera. In total, the dataset contains 1716 video clips recorded over around 77 hours with 48 distinct actions. On average, there are 6.9 action segments per video. \textbf{50 Salads} contains videos preparing 50 different salads and is recorded through an overhead Kinect camera. Videos average 6 minutes in length, and 20 segments per video. Compared to these datasets, \textbf{EgoProceL} contains a diverse set of tasks that are performed in different environment settings (e.g. assembling furniture, repairing cars, etc.). Overall, the dataset contains 1055 videos with 130 unique actions, with videos including 21 action segments on average. 
\newline

\subsection{Evaluation Metrics}

Following earlier studies on action segmentation \cite{farha2019ms, lu2024fact}, we report frame-wise accuracy (Acc) and segmentation metrics such as segmental edit distance (Edit), segmental F1 scores at 10\%, 25\%, and 50\% overlaps (F1@{10, 25, 50}). Following \cite{liu2023diffusion}, we also report the average score (Avg) across five metrics (i.e. F1@{10, 25, 50}, Edit, and Acc) as a single compact value summarising the overall frame-wise and segmentation quality of the model.

\subsection{Implementation Details}

As inputs to the frame stream of the DSA\_Net, we used I3D features \cite{carreira2017quo}, where the frame-wise feature dimensionality $d_f=2048$. The action token feature dimension, $d_a=64$, while we follow a similar approach to \cite{lu2024fact} by maintaining a fixed action token length ($M$), which we determined experimentally. 

The TE block directly following the input acts as a feature encoder that reduces the initial frame feature dimension ($d_f$) from 2048 to 64. However, for the TE blocks within the TSA module, we set the feature input and output feature dimensionality to 64. The number of TSA blocks, N=3, is decided experimentally for each dataset. The quantum circuit within the TC block is designed with $n_q=4$, and within the TC block, we repeated the number of Q-ActGM layers ($n_{ql}$), where the values for $n_{ql}=3$ and $n_q=4$ are derived experimentally.          
For all experiments, we use the Adam optimiser with a learning rate of 0.0001. We implemented classical deep learning components using the PyTorch framework \cite{pytorch}, and quantum components using the PennyLane \cite{bergholm2018pennylane} library that integrates quantum computing with machine learning, thus enabling hybrid quantum-classical computation. 

\subsection{Comparison to State-of-the-Art}

In Tables \ref{tab:gtea_results}–\ref{tab:egoprocel_results}, we report evaluation results across four datasets and compare them with state-of-the-art methods. For all datasets, our proposed DSA\_Net outperforms existing approaches by a significant margin in both frame-wise accuracy and segmentation metrics.

Regarding frame-wise accuracy (Acc), we observe consistent improvements across all datasets, ranging from 1.4\% to 2.4\%. Our method also delivers notable gains in segmentation performance: the Edit score improves by approximately 0.7\% for the GTEA and Breakfast datasets, and by 2.6\% and 3.0\% for the EgoProceL and 50Salads datasets, respectively.

Compared to the initial dual-branch based approach proposed in \cite{lu2024fact}, our method achieves average performance (Avg) improvements of 1.1\%, 1.2\%, and 2.0\% on the GTEA, Breakfast, and EgoProceL datasets, respectively. Even greater improvements are observed compared to \cite{liu2023diffusion}, with gains of 2.4\%, 2.3\%, 2.6\%, and 3.1\% on the GTEA, Breakfast, 50Salads, and EgoProceL datasets, respectively.

These improvements highlight the overall effectiveness of our proposed approach for action segmentation. To further validate the contributions of each component in DSA\_Net, we conduct ablation experiments, as detailed in Sec. \ref{sec:ablations}.

%-----------

\begin{table}[t]
\centering
\renewcommand{\arraystretch}{1.2}
\rowcolors{2}{gray!5}{white}
\setlength{\tabcolsep}{4.5pt} % Reduced spacing between columns

\resizebox{0.99\columnwidth}{!}{%
\begin{tabular}{lcccccc}
\toprule
\textbf{Method} & \textbf{F1@10} & \textbf{F1@25} & \textbf{F1@50} & \textbf{Edit} & \textbf{Acc} & \textbf{Avg} \\
\midrule
ED-TCN~\cite{lea2017temporal}       & 72.2 & 69.3 & 56.0 & 64.0 & -    & - \\
TDRN~\cite{lei2018temporal}         & 79.2 & 74.4 & 62.7 & 74.1 & 70.1 & 72.1 \\
SSA-GAN~\cite{gammulle2020fine}     & 80.6 & 79.1 & 74.2 & 76.0 & 43.3 & 70.6 \\
Bridge-Prompt~\cite{li2022bridge}   & 94.1 & 92.0 & 83.0 & 91.6 & 81.2 & 88.4 \\
MSTCN~\cite{farha2019ms}            & 87.5 & 85.4 & 74.6 & 81.4 & 79.2 & 81.6 \\
MSTCN++~\cite{li2020ms}             & 88.8 & 85.7 & 76.0 & 83.5 & 80.1 & 82.8 \\
ASRF~\cite{ishikawa2021alleviating} & 89.4 & 87.8 & 79.8 & 83.7 & 77.3 & 83.6 \\
HASR~\cite{ahn2021refining}         & 90.9 & 88.6 & 76.4 & 87.5 & 77.4 & 84.2 \\
ASFormer~\cite{yi2021asformer}      & 90.1 & 88.8 & 79.2 & 84.6 & 79.7 & 84.5 \\
MVGA~\cite{aziere2023markov}        & 91.3 & 90.0 & 79.3 & 86.4 & 80.3 & 85.5 \\
TCTr~\cite{aziere2022multistage}    & 91.3 & 90.1 & 80.0 & 87.9 & 81.1 & 86.1 \\
UVAST~\cite{behrmann2022unified}    & 92.7 & 91.3 & 81.0 & 92.1 & 80.2 & 87.5 \\
RTK~\cite{jiang2023video}           & 91.2 & 90.6 & 83.4 & 87.9 & 80.3 & 86.7 \\
DiffAct~\cite{liu2023diffusion}     & 92.5 & 91.5 & 84.7 & 89.6 & 82.2 & 88.1 \\
FACT~\cite{lu2024fact}              & 93.5 & 92.1 & 84.1 & 91.4 & 86.1 & 89.4 \\
\rowcolor{gray!20}
\textbf{DSA\_Net (Ours)}            & \textbf{94.2} & \textbf{92.8} & \textbf{85.2} & \textbf{92.1} & \textbf{88.3} & \textbf{90.5} \\
\bottomrule
\end{tabular}
}
\caption{Action segmentation results on the GTEA dataset. The best results are highlighted in bold.}
\label{tab:gtea_results}
\end{table}

%-----------------------------

\begin{table}[t]
\centering
\renewcommand{\arraystretch}{1.2}
\rowcolors{2}{gray!5}{white}
\setlength{\tabcolsep}{4.5pt} % Reduced spacing between columns

\resizebox{0.99\columnwidth}{!}{%
\begin{tabular}{lcccccc}
\toprule
\textbf{Method} & \textbf{F1@10} & \textbf{F1@25} & \textbf{F1@50} & \textbf{Edit} & \textbf{Acc} & \textbf{Avg} \\
\midrule
SSA-GAN~\cite{gammulle2020fine}       & -    & -    & -    & -    & 43.3 &  \\
MSTCN~\cite{farha2019ms}              & 52.6 & 48.1 & 37.9 & 61.7 & 63.3 & 52.7 \\
MSTCN++~\cite{li2020ms}               & 64.1 & 58.6 & 45.9 & 64.9 & 67.6 & 60.2 \\
MuCon~\cite{souri2021fast}           & 73.2 & 66.1 & 48.4 & 76.3 & 62.8 & 65.4 \\
ASRF~\cite{ishikawa2021alleviating}  & 74.3 & 68.9 & 56.1 & 72.4 & 67.6 & 67.9 \\
HASR~\cite{ahn2021refining}          & 74.7 & 69.5 & 57.0 & 71.9 & 69.6 & 68.5 \\
ASFormer~\cite{yi2021asformer}       & 76.0 & 70.6 & 57.4 & 75.0 & 73.5 & 70.5 \\
DTL~\cite{xu2022don}                 & 78.8 & 74.5 & 62.9 & 77.7 & 75.4 & 73.9 \\
MVGA~\cite{aziere2023markov}         & 75.6 & 72.1 & 59.7 & 76.8 & 72.3 & 71.3 \\
TCTr~\cite{aziere2022multistage}     & 76.6 & 71.1 & 58.5 & 76.1 & 77.5 & 72.0 \\
UVAST~\cite{behrmann2022unified}     & 76.9 & 71.5 & 58.0 & 77.1 & 69.7 & 70.6 \\
RTK~\cite{jiang2023video}            & 76.9 & 72.4 & 60.5 & 76.1 & 73.2 & 71.8 \\
LTContext~\cite{bahrami2023much}     & 77.6 & 72.6 & 60.1 & 77.0 & 74.2 & 72.3 \\
DiffAct~\cite{liu2023diffusion}      & 80.3 & 75.9 & 64.6 & 78.4 & 76.4 & 75.1 \\
FACT~\cite{lu2024fact}               & 81.4 & 76.5 & 66.2 & 79.7 & 76.2 & 76.0 \\
\rowcolor{gray!20}
\textbf{DSA\_Net (Ours)}             & \textbf{82.0} & \textbf{77.7} & \textbf{68.1} & \textbf{80.4} & \textbf{78.0} & \textbf{77.2} \\
\bottomrule
\end{tabular}
}
\caption{Action segmentation results on the Breakfast dataset. The best results are highlighted in bold.}
\label{tab:breakfast_results}
\end{table}

% ----------------------------------------------------

\begin{table}[t]
\centering
\renewcommand{\arraystretch}{1.2}
\rowcolors{2}{gray!5}{white}
\setlength{\tabcolsep}{5pt}

\resizebox{0.99\columnwidth}{!}{%
\begin{tabular}{lcccccc}
\toprule
\textbf{Method} & \textbf{F1@10} & \textbf{F1@25} & \textbf{F1@50} & \textbf{Edit} & \textbf{Acc} & \textbf{Avg} \\
\midrule
MS-TCN++~\cite{li2020ms}         & 80.7 & 78.5 & 70.1 & 74.3 & 83.7 & 77.5 \\
SSTDA~\cite{chen2020action}      & 83.0 & 81.5 & 73.8 & 75.8 & 83.2 & 79.5 \\
GTRM~\cite{huang2020improving}   & 75.4 & 72.8 & 63.9 & 67.5 & 82.6 & 72.4 \\
BCN~\cite{wang2020boundary}      & 82.3 & 81.3 & 74.0 & 74.3 & 84.4 & 79.3 \\
MTDA~\cite{chen2020action}       & 82.0 & 80.1 & 72.5 & 75.2 & 83.2 & 78.6 \\
G2L~\cite{gao2021global2local}   & 80.3 & 78.0 & 69.8 & 73.4 & 82.2 & 76.7 \\
HASR~\cite{ahn2021refining}      & 86.6 & 85.7 & 78.5 & 81.0 & 83.9 & 83.1 \\
ASRF~\cite{ishikawa2021alleviating} & 84.9 & 83.5 & 77.3 & 79.3 & 84.5 & 81.9 \\
ASFormer~\cite{yi2021asformer}   & 85.1 & 83.4 & 76.0 & 79.6 & 85.6 & 81.9 \\
UARL~\cite{chen2022uncertainty}  & 85.3 & 83.5 & 77.8 & 78.2 & 84.1 & 81.8 \\
DPRN~\cite{park2022maximization} & 87.8 & 86.3 & 79.4 & 82.0 & 87.2 & 84.5 \\
SEDT~\cite{kim2022stacked}       & 89.9 & 88.7 & 81.1 & 84.7 & 86.5 & 86.2 \\
TCTr~\cite{aziere2022multistage} & 87.5 & 86.1 & 80.2 & 83.4 & 86.6 & 84.8 \\
FAMMSDTN~\cite{du2023dilated}    & 86.2 & 84.4 & 77.9 & 79.9 & 86.4 & 82.9 \\
DTL~\cite{xu2022don}             & 87.1 & 85.7 & 78.5 & 80.5 & 86.9 & 83.7 \\
UVAST~\cite{behrmann2022unified} & 89.1 & 87.6 & 81.7 & 83.9 & 87.4 & 85.9 \\
DiffAct~\cite{liu2023diffusion}  & 90.1 & 89.2 & 83.7 & 85.0 & 88.9 & 87.4 \\
\rowcolor{gray!20}
\textbf{DSA\_Net (Ours)}         & \textbf{92.7} & \textbf{92.3} & \textbf{87.1} & \textbf{88.8} & \textbf{91.3} & \textbf{90.4} \\
\bottomrule
\end{tabular}
}
\caption{Action segmentation results on the 50Salads dataset. The best results are highlighted in bold.}
\label{tab:50salads_results}
\end{table}

%-----------------------------------------------------

\begin{table}[t]
\centering
\renewcommand{\arraystretch}{1.2}
\rowcolors{2}{gray!5}{white}
\setlength{\tabcolsep}{4.5pt} % Reduced spacing between columns

\resizebox{0.95\columnwidth}{!}{%
\begin{tabular}{lccccccc}
\toprule
\textbf{Method} & \textbf{F1@10} & \textbf{F1@25} & \textbf{F1@50} & \textbf{Edit} & \textbf{Acc} & \textbf{Avg} & \textbf{AccB} \\
\midrule
MSTCN++~\cite{li2020ms}           & 60.3 & 57.0 & 46.5 & 62.4 & 69.3 & 59.1 & 82.5 \\
ASFormer~\cite{yi2021asformer}    & 63.3 & 60.9 & 51.0 & 64.9 & 71.1 & 62.2 & 84.9 \\
UVAST~\cite{behrmann2022unified}  & 60.5 & 58.3 & 46.6 & 67.7 & 67.8 & 60.2 & 83.2 \\
LTContext~\cite{bahrami2023much}  & 64.2 & 61.3 & 51.2 & 61.3 & 70.3 & 61.7 & 84.7 \\
DiffAct~\cite{liu2023diffusion}   & 67.5 & 65.4 & 54.6 & 68.4 & 77.0 & 66.6 & 86.6 \\
FACT~\cite{lu2024fact}            & 73.0 & 69.8 & 60.8 & 75.7 & 77.6 & 71.4 & 88.0 \\
\rowcolor{gray!20}
\textbf{DSA\_Net (Ours)}          & \textbf{75.1} & \textbf{72.6} & \textbf{62.1} & \textbf{78.3} & \textbf{79.0} & \textbf{73.4} & \textbf{89.7} \\
\bottomrule
\end{tabular}
}
\caption{Action segmentation results on the EgoProceL dataset. The best results are highlighted in bold.}
\label{tab:egoprocel_results}
\end{table}

\subsection{Ablation Experiments}
\label{sec:ablations}

We performed a series of ablation experiments to systematically evaluate the contribution of each innovation proposed through our DSA\_Net framework. The following sections discuss the effect of adding each of the key innovations.

\subsubsection{Effect of the number of TSA Modules}

As discussed in Sec. \ref{sec:methods}, our proposed TSA block integrates temporal encoding with the Q-ActGM fusion mechanism. Table \ref{tab:n_tsa} presents the impact of varying the number of TSA blocks (denoted as N) on action segmentation performance. The model achieved its best results on the 50Salads dataset with N = 3, and beyond this point, performance begins to decline.

This suggests that stacking TSA blocks facilitates progressive feature refinement, enhancing the model’s ability to capture spatio-temporal dependencies crucial for accurate action classification and segmentation. Notably, even with a single TSA block, the model achieves considerable accuracy, indicating the effectiveness of the proposed approach. However, increasing the number of blocks helps reduce misclassifications and over-segmentation errors. These findings suggest that a moderate number of TSA blocks is sufficient to balance performance and model complexity.

\begin{table}[t]
\centering
\renewcommand{\arraystretch}{1.2}
\setlength{\tabcolsep}{6pt}

\begin{tabular}{lcccc}
\toprule
N  & \textbf{F1@{10, 25, 50}} & \textbf{Edit} & \textbf{Acc} \\
\midrule
1 & 74.2, 70.7, 68.1   & 72.3 & 76.0 \\
2 &  89.8, 89.4, 82.2 & 85.3  & 88.7 \\
3 & \textbf{92.7}, \textbf{92.3}, \textbf{87.1} & \textbf{88.8} & \textbf{91.3} \\
4 &  92.3, 92.2, 87.2  & 88.6 & 91.1 \\
5 &  91.8, 91.7, 85.8  & 87.0 & 89.9 \\

\bottomrule
\end{tabular}
\caption{Effect of the number of TSA blocks ($N$) on the 50Salads dataset.}
\label{tab:n_tsa}
\end{table}

\subsubsection{Effect of Training Losses}

In the proposed work, the dual-stream alignment loss plays a critical role in guiding the frame and action branches to learn action segmentation-relevant features through feature distribution alignment. As discussed, this loss comprises three components: a relational consistency loss ($\mathcal{L}{rel}$), a cross-level contrastive ($\mathcal{L}{clc}$) loss, and a cycle-consistency reconstruction loss ($\mathcal{L}_{cyc}$).

Table \ref{tab:loss_ablation_50salads} presents an ablation study evaluating the impact of these components, alongside the frame-wise ($\mathcal{L}{ce_f}$) and action-wise ($\mathcal{L}{ce_a}$) cross-entropy losses. The model already achieves strong performance using only the cross-entropy losses. However, the addition of each alignment loss component leads to consistent and significant improvements.

Specifically, incorporating $\mathcal{L}{rel}$, $\mathcal{L}{clc}$, and $\mathcal{L}_{cyc}$ results in accuracy gains of 4.9\%, 5.5\%, and 5.7\%, respectively, over the baseline. Corresponding improvements in the Edit score are 3.8\%, 5.0\%, and 5.4\%, respectively. These results highlight the effectiveness of the dual-stream alignment loss in enhancing the model’s ability to capture spatio-temporal relationships for accurate action segmentation.

\begin{table}[t]
\centering
\begin{tabular}{ccccc|ccc}
\toprule
A & B & C & D & E & \textbf{F1@\{10, 25, 50\}} & \textbf{Edit} & \textbf{Acc} \\
\midrule
\checkmark & & & & & 84.2, 83.9, 78.8 & 80.1 & 84.7\\
\checkmark & \checkmark & & & & 86.9, 85.7, 80.2  & 83.4 & 85.6\\
\checkmark & \checkmark & \checkmark & & & 90.7, 90.5, 85.2 & 87.2& 90.5\\
\checkmark & \checkmark & \checkmark & \checkmark & & 91.5, 91.1, 86.9 & 88.4& 91.1\\
\checkmark & \checkmark & \checkmark & \checkmark & \checkmark & \textbf{92.7, 92.3, 87.1} & \textbf{88.8} & \textbf{91.3} \\

\bottomrule
\end{tabular}
\caption{Ablation study considering the five loss terms. Here A, B, C, D and E represent the cross-entropy losses corresponding to frame branch ($\mathcal{L}_{ce_{f}}$) and action branch ($\mathcal{L}_{ce_{a}}$), and the 3 loss components of the dual-stream alignment loss: relational consistency ($\mathcal{L}_{rel}$), cross-level contrastive ($\mathcal{L}_{clc}$) and cycle-consistency reconstruction ($\mathcal{L}_{cyc}$) losses. Experiments are performed using the 50 Salads dataset.}
\label{tab:loss_ablation_50salads}
\end{table}

\subsubsection{Effect of Q-ActGM}
\label{sec:q_actgm}

In the Temporal Context (TC) block, we adopt the concept of feature modulation via the ActGM layer to fuse action-wise tokens with frame-wise features. To further enhance the expressive capacity of the TC block, we introduce quantum properties through the proposed Q-ActGM layer. Table \ref{tab:q_actgm} reports results for an ablation study comparing the fusion with and without integrating the quantum properties in the proposed ActGM module (i.e. ActGM vs Q-ActGM).

The results demonstrate that the Q-ActGM layer significantly improves both frame-wise and segmentation performance. Specifically, Q-ActGM achieves an improvement of 2.6\% in accuracy and 2.5\% in Edit score over ActGM, highlighting its superior expressive power in capturing spatio-temporal dependencies for action segmentation.

\begin{table}[t]
\centering
\renewcommand{\arraystretch}{1.2}
\setlength{\tabcolsep}{6pt}

\begin{tabular}{lcccc}
\toprule
TC Block & F1@{10, 25, 50} & Edit & Acc\\
\midrule
with ActGM &  89.6, 88.9, 83.3  & 86.3 & 88.7 \\
with Q-ActGM &  \textbf{92.7, 92.3, 87.1}  & \textbf{88.8} & \textbf{91.3}\\

\bottomrule
\end{tabular}
\caption{Effect of the quantum-based feature modulation (Q-ActGM) on 50Salads dataset.}
\label{tab:q_actgm}
\end{table}

\subsubsection{Effect of Quantum-based Hyperparameters} 

As discussed earlier, features are projected to a number of qubits ($n_q$) before being passed through the Q-ActGM layer. Within the Q-ActGM circuit (see Sec. \ref{para:q_circuit}), the entangling layers are repeated $n_{ql}$ times. Table \ref{tab:quantum_ablation} presents results for various combinations of $n_q$ and $n_{ql}$. Our experiments indicate that the hybrid quantum-classical model achieves its best performance when $n_q = 3$ and $n_{ql} = 3$.

The results suggest that moderately deep circuits with a limited number of qubits are sufficient to effectively model temporal action features during the fusion process. We believe this configuration reflects a balance between expressivity and resource efficiency.

\begin{table}[htbp!]
\centering
\renewcommand{\arraystretch}{1.2}
\rowcolors{2}{gray!5}{white}
\setlength{\tabcolsep}{5pt}

\resizebox{0.85\columnwidth}{!}{%
\begin{tabular}{ccccccc}
\toprule
\textbf{$n_{q}$} & \textbf{$n_{ql}$} & \textbf{F1@10} & \textbf{F1@25} & \textbf{F1@50} & \textbf{Edit} & \textbf{Acc} \\
\midrule
2 & 1 & 83.9 & 80.1 & 68.9 & 80.0 & 86.1 \\
2 & 3 & 82.0 & 77.3 & 68.0 & 78.2 & 87.9 \\
2 & 5 & 83.5 & 80.2 & 69.0 & 77.6 & 87.9 \\
3 & 1 & 89.3 & 83.2 & 70.1 & 83.7 & 88.8 \\
\textbf{3} & \textbf{3} & \textbf{92.7} & \textbf{92.3} & \textbf{87.1} & \textbf{88.8} & \textbf{91.3} \\
3 & 5 & 90.3 & 87.9 & 85.5 & 85.8 & 84.1 \\
4 & 1 & 83.9 & 80.1 & 78.6 & 80.6 & 82.5 \\
4 & 3 & 79.7 & 76.8 & 63.6 & 76.1 & 82.3 \\
4 & 5 & 73.0 & 71.0 & 59.3 & 71.9 & 78.5 \\

\bottomrule
\end{tabular}
}
\caption{Effect of the quantum-based hyperparameters, $n_{q}$ and {$n_{ql}$}, on the 50Salads dataset.}
\label{tab:quantum_ablation}
\end{table}

%PennyLane \cite{bergholm2018pennylane}

\section{Qualitative Results}

\begin{figure*}[htbp!]
    \centering
    \subfloat{%
        \includegraphics[width=0.9\linewidth]{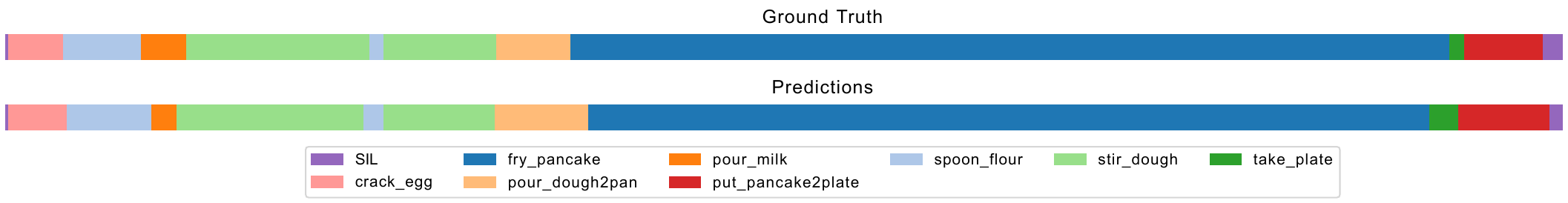}%
        \label{fig:bf_a1}%
    }
    \hfill
    \subfloat{%
        \includegraphics[width=0.9\linewidth]{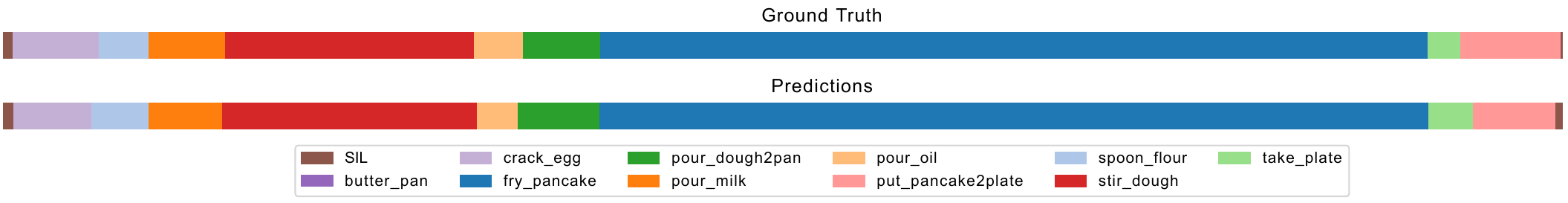}%
        \label{fig:bf_b1}%
    }
    \caption{Visualisation of the action segmentation results on the Breakfast dataset.}
    \label{fig:breakfast-timelines}
\end{figure*}

\begin{figure*}[htbp!]
    \centering
    \subfloat{%
        \includegraphics[width=0.9\linewidth]{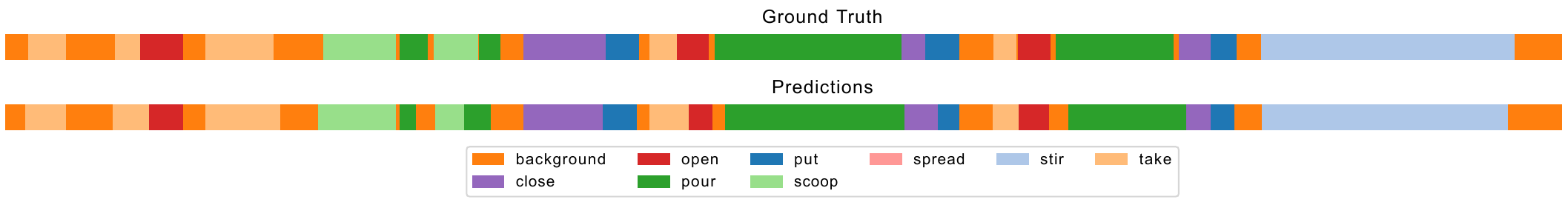}%
        \label{fig:bf_a2}%
    }
    \hfill
    \subfloat{%
        \includegraphics[width=0.9\linewidth]{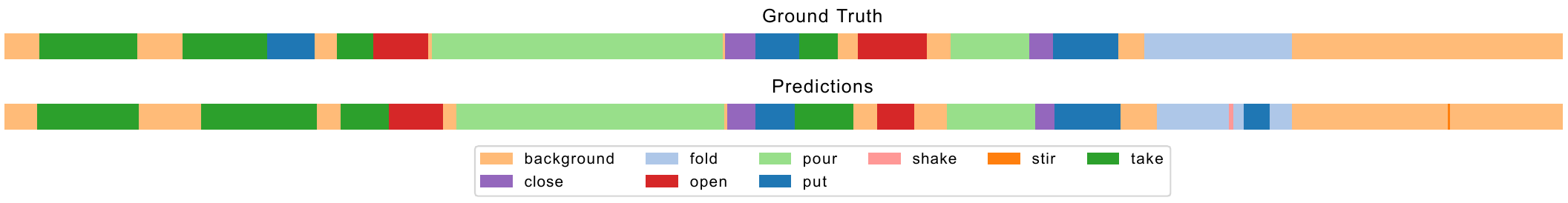}%
        \label{fig:bf_b2}%
    }
    \caption{Visualisation of the action segmentation results on the GTEA dataset.}
    \label{fig:gtea-timelines}
\end{figure*}

\begin{figure*}[htbp!]
    \centering
    \subfloat{%
        \includegraphics[width=0.9\linewidth]{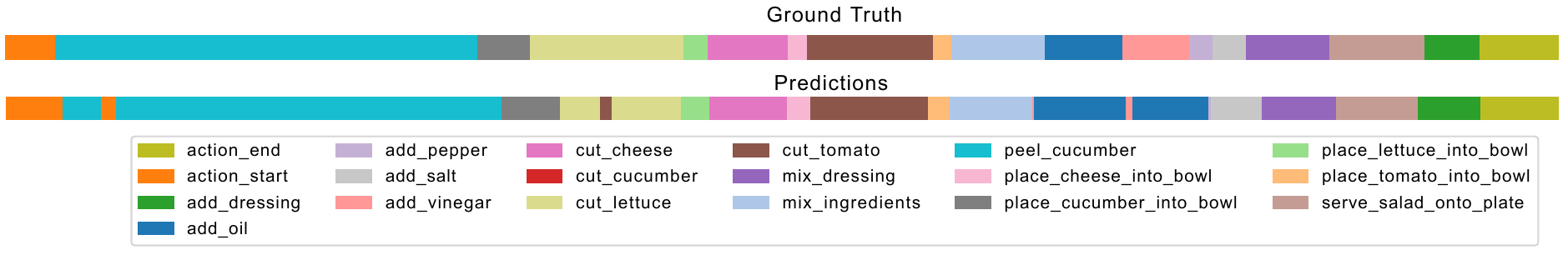}%
        \label{fig:bf_a3}%
    }
    \hfill
    \subfloat{%
        \includegraphics[width=0.9\linewidth]{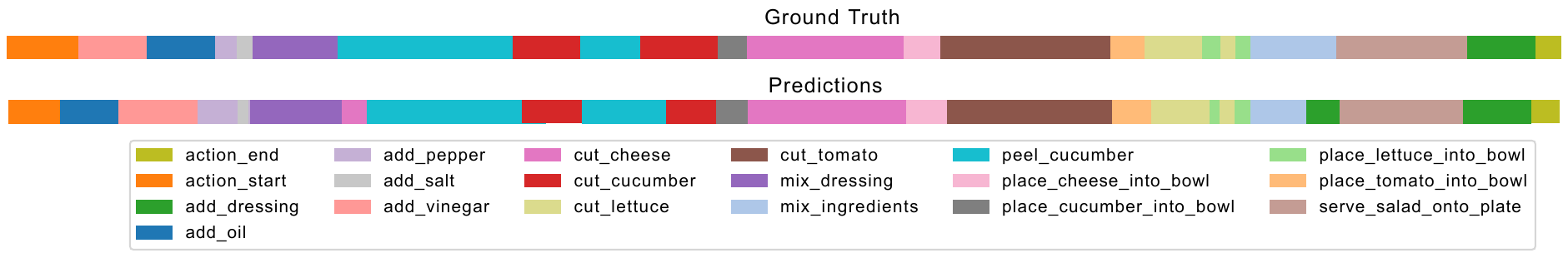}%
        \label{fig:bf_b3}%
    }
    \caption{Visualisation of the action segmentation results on the 50 Salads dataset.}
    \label{fig:50salads-timelines}
\end{figure*}

\begin{figure*}[htbp!]
    \centering
    \subfloat{%
        \includegraphics[width=0.9\linewidth]{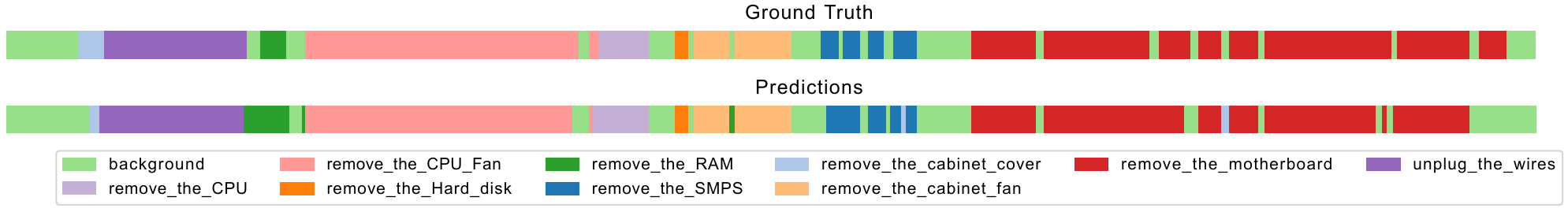}%
        \label{fig:bf_a4}%
    }
    \hfill
    \subfloat{%
        \includegraphics[width=0.9\linewidth]{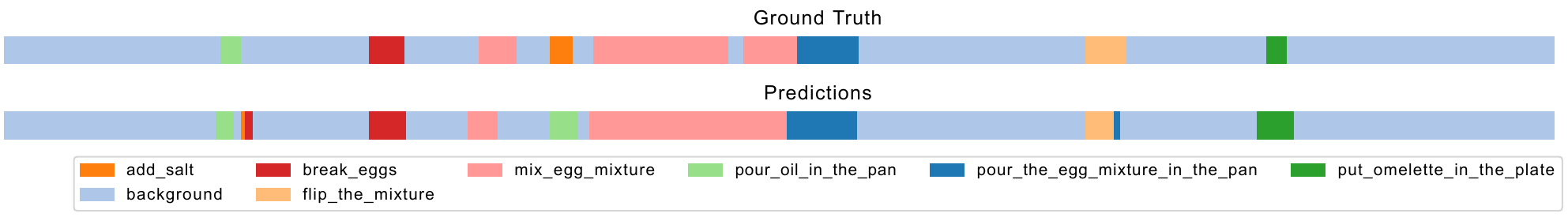}%
        \label{fig:bf_b4}%
    }
    \caption{Visualisation of the action segmentation results on the EgoProceL dataset.}
    \label{fig:egoprocel-timelines}
\end{figure*}

In this section, we further illustrate the performance of the proposed DSA\_Net with qualitative results. In Figures \ref{fig:breakfast-timelines}, \ref{fig:gtea-timelines}, \ref{fig:50salads-timelines}, \ref{fig:egoprocel-timelines}, we visualise and compare the predictions obtained for the four datasets with their corresponding ground truth annotations. 

Across all datasets, we observe occasional discrepancies in the timing of action transitions, where predicted transitions are either slightly early or delayed compared to the ground truth. For instance, in the Breakfast dataset (see Figure~\ref{fig:breakfast-timelines}), transitions such as \textit{fry\_pancake}~$\rightarrow$~\textit{take\_plate} and \textit{pour\_milk}~$\rightarrow$~\textit{stir\_dough} were predicted slightly earlier than the ground truth, whereas transitions like \textit{spoon\_flour}~$\rightarrow$~\textit{pour\_milk} and \textit{take\_plate}~$\rightarrow$~\textit{put\_pancake2plate} were delayed.

In the GTEA dataset, the proposed DSA\_Net demonstrated strong performance in segmenting actions, even with frequent action transitions. However, minor confusion was noted between background frames (non-action segments) and action classes (see Figure~\ref{fig:gtea-timelines}). For the 50Salads and EgoProceL datasets, some action misclassifications were observed. For example, in 50Salads, the model briefly predicted \textit{cut\_tomato} while the actual action \textit{cut\_lettuce} was being performed (see top timeline in Figure~\ref{fig:50salads-timelines}). Nevertheless, the model was able to quickly correct these errors and continued with accurate predictions. A similar pattern was observed in EgoProceL, where during \textit{remove\_the\_SMPS}, the model briefly predicted \textit{remove\_the\_cabinet\_cover}, but corrected itself shortly thereafter. Additionally, in EgoProceL, background frames were occasionally misclassified as actions such as \textit{remove\_the\_RAM}, \textit{remove\_the\_cabinet\_cover}, \textit{break\_eggs}, or \textit{pour\_the\_egg\_mixture} (see Figure~\ref{fig:egoprocel-timelines}).

Despite these occasional misclassifications, \textbf{DSA\_Net} consistently demonstrated robust performance across all four datasets, achieving significant improvements in action segmentation results and outperforming existing state-of-the-art methods.

% \begin{figure*}[htbp!]
%     \centering
%     \begin{subfigure}
%         \centering
%         \includegraphics[width=\linewidth]{Figures/timelines/bf_P14_webcam01_P14_pancake_plot.pdf}
%         % \caption{P14 sequence}
%         \label{fig:bf_a}
%     \end{subfigure}
%     \hfill
%     \begin{subfigure}
%         \centering
%         \includegraphics[width=\linewidth]{Figures/timelines/bf_P15_webcam01_P15_pancake_plot.pdf}
%         % \caption{P15 sequence}
%         \label{fig:bf_b}
%     \end{subfigure}    
%     \caption{Visualisation of the action segmentation results on the Breakfast dataset.}
%     \label{fig:breakfast-timelines}
% \end{figure*}

% \begin{figure*}[htbp!]
%     \centering
%     \begin{subfigure}
%         \centering
%         \includegraphics[width=\linewidth]{Figures/timelines/gtea_S1_CofHoney_C1_plot.pdf}
%         % \caption{P14 sequence}
%         \label{fig:bf_a}
%     \end{subfigure}
%     \hfill
%     \begin{subfigure}
%         \centering
%         \includegraphics[width=\linewidth]{Figures/timelines/gtea_S1_Hotdog_C1_plot.pdf}
%         % \caption{P15 sequence}
%         \label{fig:bf_b}
%     \end{subfigure}    
%     \caption{Visualisation of the action segmentation results on the GTEA dataset.}
%     \label{fig:gtea-timelines}
% \end{figure*}

% \begin{figure*}[htbp!]
%     \centering
%     \begin{subfigure}
%         \centering
%         \includegraphics[width=\linewidth]{Figures/timelines/50_salads_rgb-03-1_plot.pdf}
%         % \caption{P14 sequence}
%         \label{fig:bf_a}
%     \end{subfigure}
%     \hfill
%     \begin{subfigure}
%         \centering
%         \includegraphics[width=\linewidth]{Figures/timelines/50_salads_rgb-04-1_plot.pdf}
%         % \caption{P15 sequence}
%         \label{fig:bf_b}
%     \end{subfigure}    
%     \caption{Visualisation of the action segmentation results on the 50 Salads dataset.}
%     \label{fig:50salads-timelines}
% \end{figure*}

% \begin{figure*}[htbp!]
%     \centering
%     \begin{subfigure}
%         \centering
%         \includegraphics[width=\linewidth]{Figures/timelines/egoP_Head_27_plot.pdf}
%         % \caption{P14 sequence}
%         \label{fig:bf_a}
%     \end{subfigure}
%     \hfill
%     \begin{subfigure}
%         \centering
%         \includegraphics[width=\linewidth]{Figures/timelines/egoP_S09_Eggs_7150991-1050_plot.pdf}
%         % \caption{P15 sequence}
%         \label{fig:bf_b}
%     \end{subfigure}    
%     \caption{Visualisation of the action segmentation results on the EgoProceL dataset.}
%     \label{fig:egoprocel-timelines}
% \end{figure*}

\begin{figure*}[htbp!]
    \centering
    \includegraphics[width=1.0\linewidth]{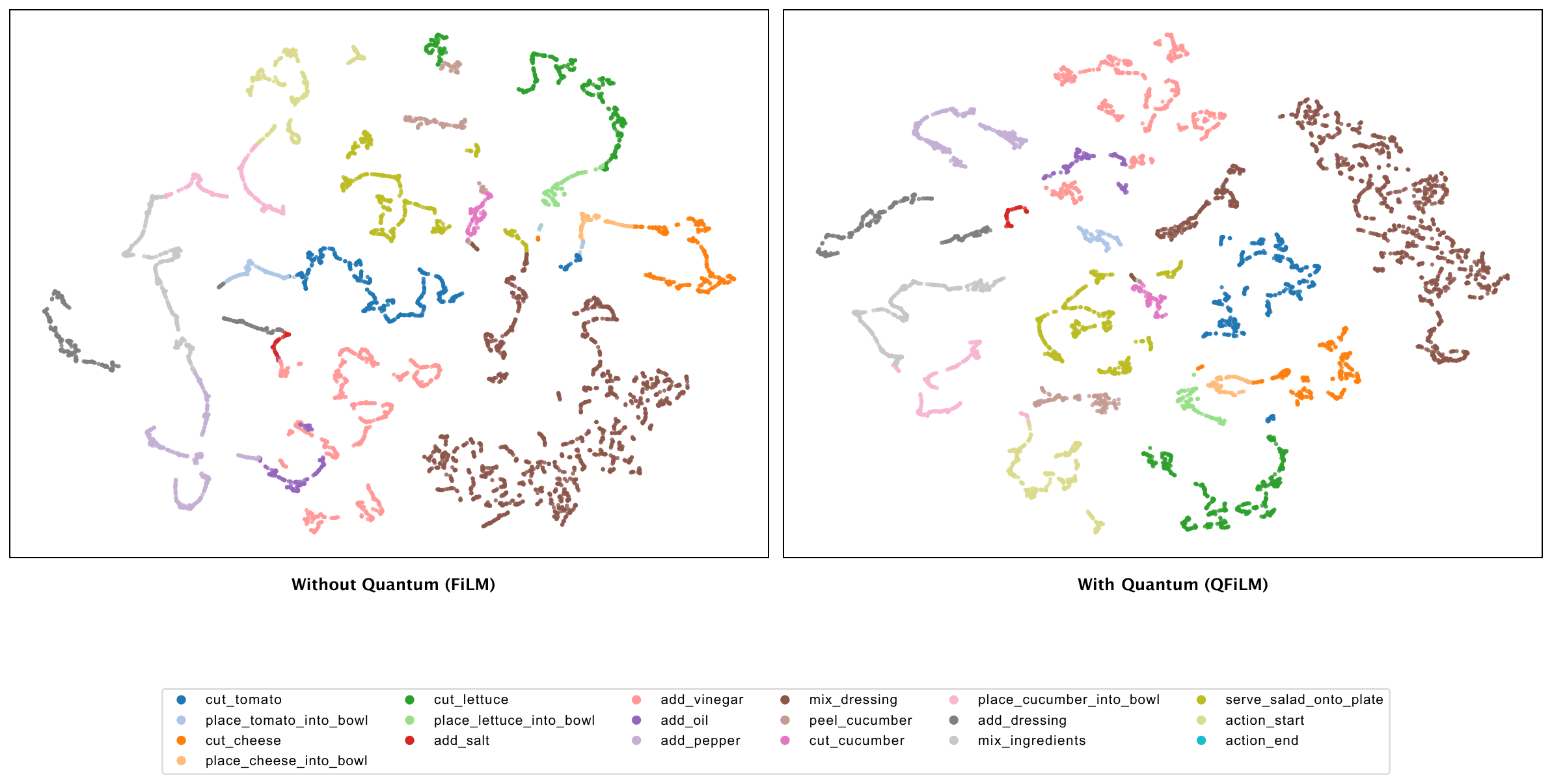}
	\caption{Visualisation of the feature embeddings using t-SNE plots using the ActGM (left) and Q-ActGM (right) formulations corresponding to Tab. \ref{tab:q_actgm}. The visualisation with Q-ActGM demonstrates improved separation and tighter clustering of several action classes, such as \textit{add\_pepper}, \textit{add\_dressing}, \textit{place\_tomato\_in\_the\_bowl}, \textit{cut\_lettuce}, and \textit{action\_end}.}
	\label{fig:tsne_outputs}
\end{figure*}

As discussed in the previous section (Sec. \ref{sec:q_actgm}), the conversion of the proposed ActGM model to its proposed quantum-based model (i.e. Q-ActGM) has significantly improved the overall action segmentation performance, where an improvement of 2.6\% in accuracy and 2.5\% in Edit score was achieved. To further provide a deeper understanding of the learned feature representations, in Fig. \ref{fig:tsne_outputs} we visualise the feature embeddings using t-SNE plots for the experiments with and without the quantum-based feature modulation, providing a qualitative comparison of the clustering behaviour and class separability achieved by the models. Without quantum feature modulation (see left sub-figure in Fig. \ref{fig:tsne_outputs}), the clusters in the t-SNE visualisation appear less distinct, with noticeable overlap between semantically similar actions (e.g., \textit{add\_oil} and \textit{add\_vinegar}). In contrast, the visualisation with Q-ActGM (see right sub-figure in Fig. \ref{fig:tsne_outputs}) demonstrates improved separation and tighter clustering of several action classes, such as \textit{add\_pepper}, \textit{add\_dressing}, \textit{place\_tomato\_in\_the\_bowl}, \textit{cut\_lettuce}, and \textit{action\_end}. The qualitative evidence supports the effectiveness of Q-ActGM in improving complementary information fusion between frame-level and action-related features, leading to better temporal action segmentation.

\section{Conclusion}

In this paper, we introduced DSA\_Net, a novel framework for action segmentation that integrates frame-wise and action-wise representations through a two-stream architecture, guided by the proposed Dual-Stream Alignment Loss. To the best of our knowledge, this is the first application of a hybrid quantum-classical machine learning model in this domain, leveraging quantum properties to enhance information fusion via the Q-ActGM module. Our method achieves state-of-the-art performance across four benchmark datasets, demonstrating the effectiveness of our dual-stream design and alignment strategy. Extensive ablation studies further validate the contributions of each component.

Future work could explore avenues to improve efficiency, particularly for real-time deployment, and the application of DSA\_Net to other video understanding tasks. Moreover, this work also opens new directions towards integrating quantum principles into deep learning architectures for video analysis, setting the stage for further exploration in hybrid quantum-classical learning systems.

\section*{Acknowledgments}
This research was supported by an Australian Research Council (ARC) Discovery grant DP250103634.


\clearpage
\balance
\bibliographystyle{IEEEtran}
% \FloatBarrier
\bibliography{IEEEexample.bib}

% \section{References Section}
% You can use a bibliography generated by BibTeX as a .bbl file.
%  BibTeX documentation can be easily obtained at:
%  http://mirror.ctan.org/biblio/bibtex/contrib/doc/
%  The IEEEtran BibTeX style support page is:
%  http://www.michaelshell.org/tex/ieeetran/bibtex/
 
%  % argument is your BibTeX string definitions and bibliography database(s)
% %\bibliography{IEEEabrv,../bib/paper}
% You can manually copy in the resultant .bbl file and set second argument of $\backslash${\tt{begin}} to the number of references
%  (used to reserve space for the reference number labels box).

\newpage
% \balance
\section{Biography Section}

 

% % \begin{IEEEbiography}[{\includegraphics[width=1in,height=1.25in,clip,keepaspectratio]{fig1}}]{Michael Shell}
% % \end{IEEEbiography}

\begin{IEEEbiography}[{\includegraphics[width=1in,height=1.25in,clip,keepaspectratio]{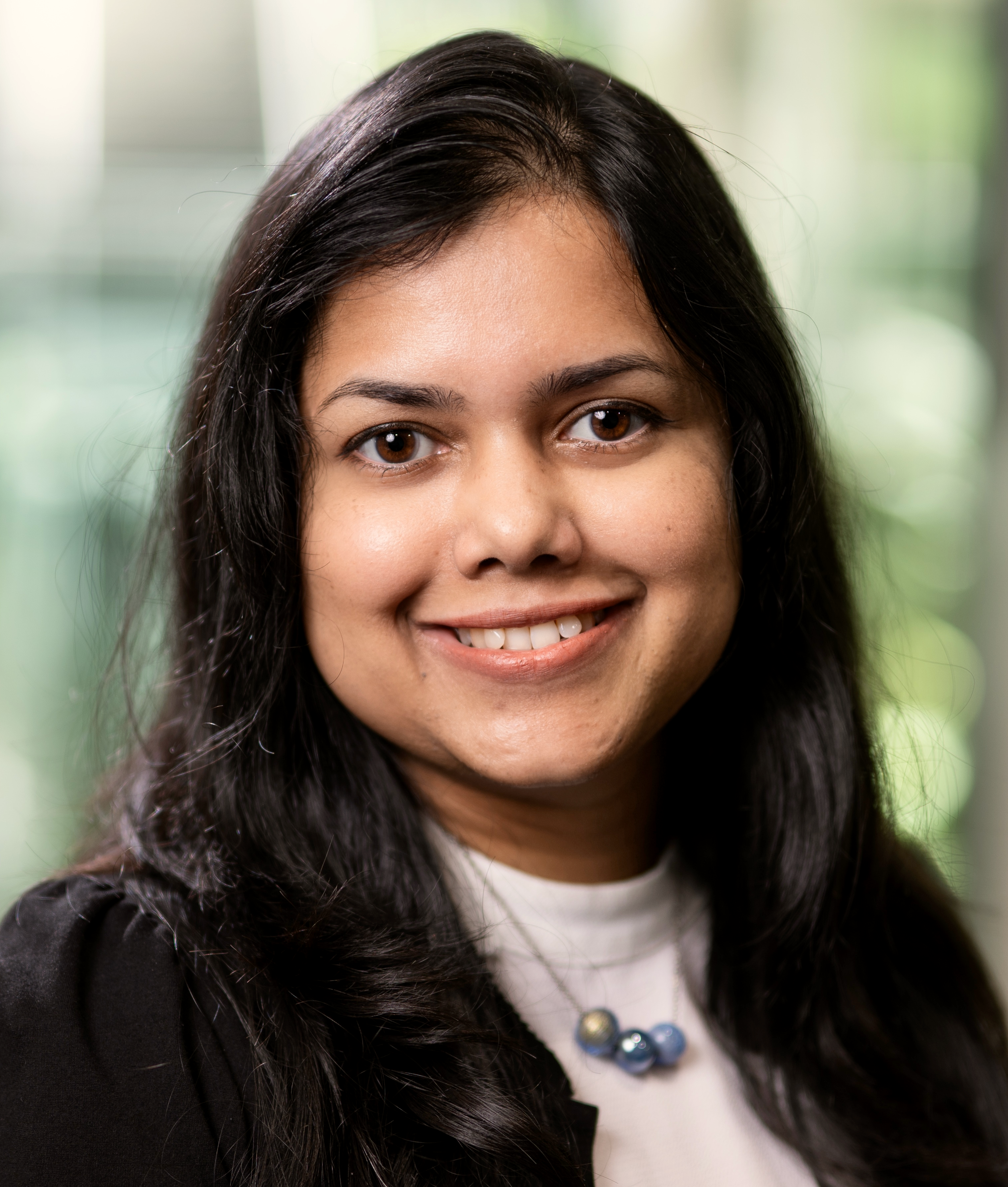}}]{Harshala Gammulle} received her BSc (Hons) in Computer Science from the University of Peradeniya, Sri Lanka, and her PhD from the Queensland University of Technology (QUT), Australia. She is currently a Postdoctoral Research Fellow in the Signal Processing, Artificial Intelligence, and Vision Technologies (SAIVT) research program within the School of Electrical Engineering and Robotics at QUT. Dr Gammulle is the recipient of the 2019 QUT Executive Dean’s Commendation for Outstanding Doctoral Thesis Award and the QUT Early Career Researcher Award in 2023. Her research expertise lies in machine learning and computer vision, with a strong focus on spatio-temporal modelling for human behaviour understanding and the development of hybrid quantum-classical machine learning models. 
\end{IEEEbiography} \vspace{-33pt}

\begin{IEEEbiography}[{\includegraphics[width=1in,height=1.25in,clip,keepaspectratio]{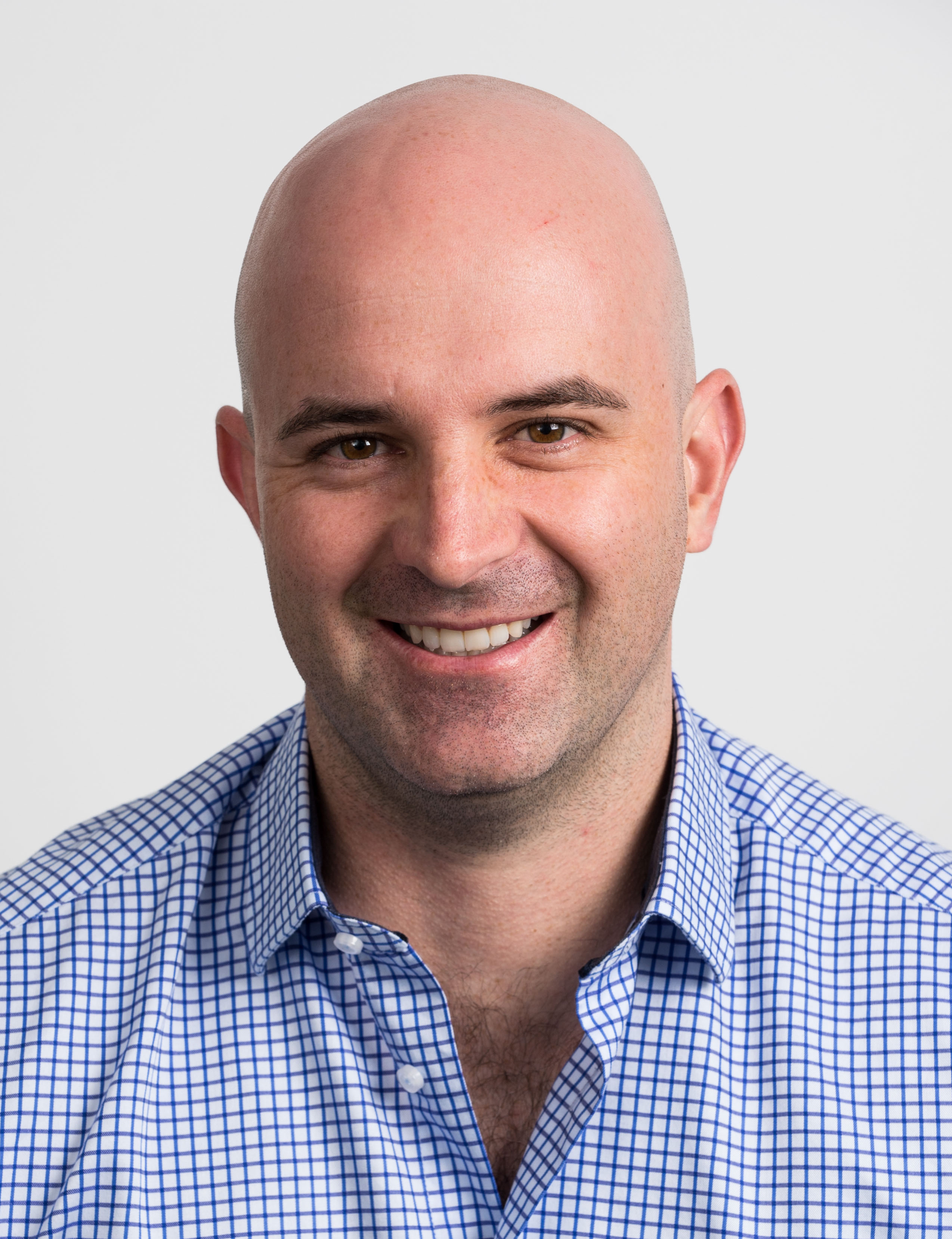}}]{Clinton Fookes}{\space} received the B.Eng. in Aerospace/Avionics, the MBA degree, and the Ph.D. degree in computer vision. He is currently the Associate Dean Research, a Professor of Vision and Signal Processing, and Co-Director of the SAIVT Lab (Signal Processing, Artificial Intelligence and Vision Technologies) with the Faculty of Engineering at the Queensland University of Technology, Brisbane, Australia. His research interests include computer vision, machine learning, signal processing, and artificial intelligence. He serves on the editorial boards for IEEE TRANSACTIONS ON IMAGE PROCESSING and Pattern Recognition. He has previously served on the Editorial Board for IEEE TRANSACTIONS ON INFORMATION FORENSICS AND SECURITY. He is a Fellow of the International Association of Pattern Recognition, a Fellow of the Australian Academy of Technological Sciences and Engineering, and a Fellow of the Asia-Pacific Artificial Intelligence Association. He is a Senior Member of the IEEE and a multi-award winning researcher including an Australian Institute of Policy and Science Young Tall Poppy, an Australian Museum Eureka Prize winner, Engineers Australia Engineering Excellence Award, Australian Defence Scientist of the Year, and a Senior Fulbright Scholar.
\end{IEEEbiography} \vspace{-33pt}

\begin{IEEEbiography}[{\includegraphics[width=1in,height=1.25in,clip,keepaspectratio]{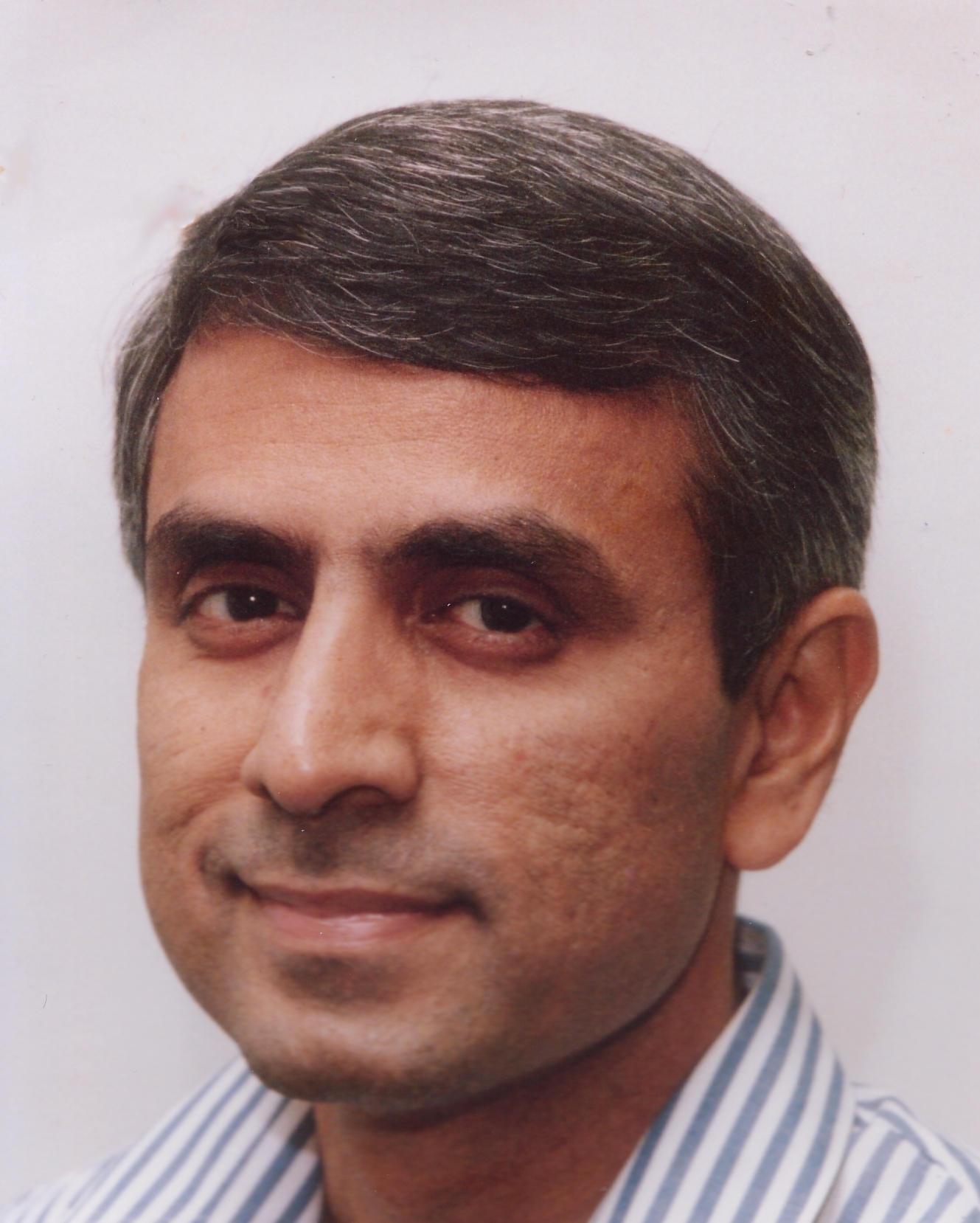}}]{Sridha Sridharan} has obtained an MSc (Communication Engineering) degree from the University of Manchester, UK, and a PhD degree from the University of New South Wales, Australia. He is currently with the Queensland University of Technology (QUT) where he is a Professor in the School of Electrical Engineering and  Robotics. He has published over 600 papers consisting of publications in journals and in refereed international conferences in the areas of Image and Speech technologies during the period 1990-2023.  During this period he has also graduated 85  PhD students in the areas of Image and Speech technologies. Prof Sridharan has also received a number of research grants from various funding bodies including the Commonwealth competitive funding schemes such as the Australian Research Council (ARC) and the National Security Science and Technology (NSST) unit. Several of his research outcomes have been commercialised.
\end{IEEEbiography} \vspace{-33pt}

\begin{IEEEbiography}[{\includegraphics[width=1in,height=1.25in,clip,keepaspectratio]{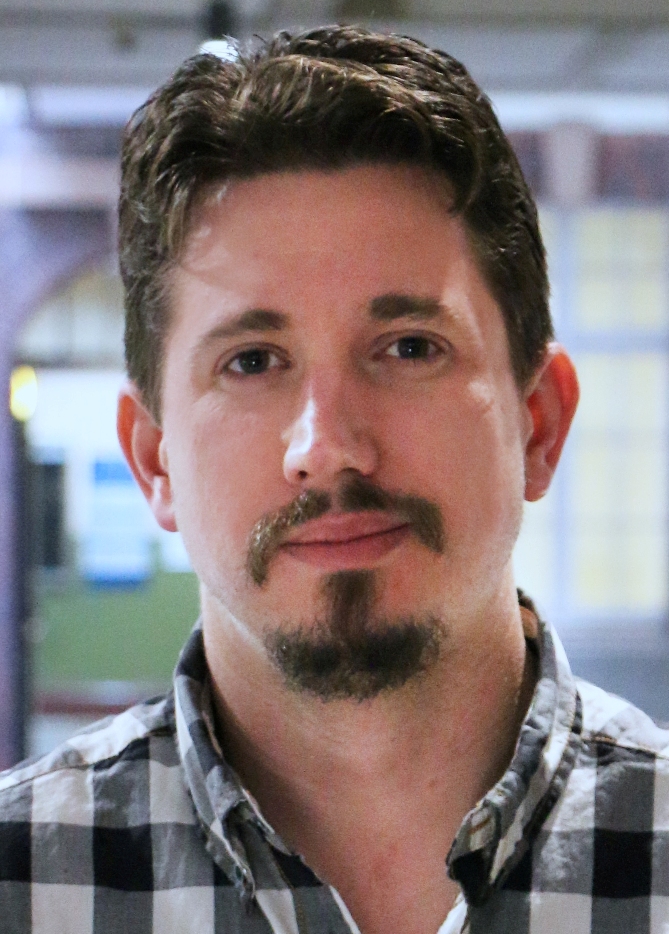}}]{Simon Denman} is an Associate Professor in the School of Electrical Engineering and Robotics at Queensland University of Technology (QUT). Simon actively researches in the fields of computer vision and machine learning, including action and event recognition, trajectory prediction, video analytics, biometrics, and medical signal processing. Simon has published over 200 papers in the areas of computer vision and machine learning, and co-leads the Applied Data Science research programme within the QUT Centre for Data Science.
\end{IEEEbiography}



\vfill

\end{document}